\documentclass[sigconf]{acmart}

\usepackage{amsmath}
\usepackage{amsthm}
\usepackage{nicefrac}
\usepackage{algorithm}
\usepackage{tcolorbox}
\usepackage{apxproof}
\usepackage{proof-at-the-end}

\newtheorem{manualtheoreminner}{Theorem}
\newenvironment{manualtheorem}[1]{%
  \renewcommand\themanualtheoreminner{#1}%
  \manualtheoreminner
}{\endmanualtheoreminner}

\newcommand{\va}{\mathbf{a}}

\newcommand{\vy}{\mathbf{y}}
\newcommand{\ba}{\mathbf{a}}
\newcommand{\bb}{\mathbf{b}}
\newcommand{\bc}{\mathbf{c}}
\newcommand{\bt}{\mathbf{t}}
\newcommand{\by}{\mathbf{y}}

\DeclareMathOperator{\E}{\mathbb{E}}
\DeclareMathOperator{\parity}{par}

\theoremstyle{definition}
\newtheorem{definition}{Definition}
\newtheorem{problem}{Problem}

\makeatletter
\def\thm@space@setup{\thm@preskip=5pt
\thm@postskip=1pt}
\makeatother
\newtheoremstyle{newstyle}      
{} 
{} 
{\itshape} 
{10pt} 
{\scshape} 
{.} 
{ } 
{} 

\theoremstyle{newstyle}
\newtheorem{thm}{Theorem}[section]

\newenvironment{pfs}{{\itshape Proof sketch.}\begin{mdseries}}{\end{mdseries}}

\usepackage{algorithm}
\usepackage[noend]{algpseudocode}

\AtBeginDocument{%
  \providecommand\BibTeX{{%
    \normalfont B\kern-0.5em{\scshape i\kern-0.25em b}\kern-0.8em\TeX}}}

\acmDOI{10.1145/3512290.3528746} 
\acmISBN{978-1-4503-9237-2/22/07} 
\acmConference[GECCO '22]{The Genetic and Evolutionary Computation Conference 2022}{July 9--13, 2022}{Boston, USA}
\acmYear{2022}
\copyrightyear{2022}
\setcopyright{acmlicensed}

\begin{document}

\title[Expressive Encodings: Universal Approximation and other Advantages]{Simple Genetic Operators are Universal\\Approximators of Probability Distributions\\(and other Advantages of Expressive Encodings)}

\author{Elliot Meyerson}
\affiliation{
\institution{Cognizant AI Labs}
\city{San Francisco}
\country{USA}
}
\email{elliot.meyerson@cognizant.com}

\author{Xin Qiu}
\affiliation{
\institution{Cognizant AI Labs}
\city{San Francisco}
\country{USA}
}
\email{xin.qiu@cognizant.com}

\author{Risto Miikkulainen}
\affiliation{
\institution{UT Austin \& Cognizant AI Labs}
\city{Austin \& San Francisco}
\country{USA}
}
\email{risto@cs.utexas.edu}

\renewcommand{\shortauthors}{Meyerson, Qiu, and Miikkulainen}

\begin{abstract}
This paper characterizes the inherent power of evolutionary algorithms.
This power depends on the computational properties of the genetic encoding.
With some encodings, two parents recombined with a simple crossover operator
can sample from an arbitrary distribution of child phenotypes.
Such encodings are termed \emph{expressive encodings} in this paper. Universal function approximators, including
popular evolutionary substrates of genetic programming and neural networks, can be used to construct expressive encodings. Remarkably, this approach need not be applied only to domains where the phenotype is a function: Expressivity can be achieved even when optimizing static structures, such as binary vectors.
Such simpler settings make it possible to characterize expressive encodings theoretically: 
Across a variety of test problems, expressive encodings are shown to achieve up to super-exponential convergence speed-ups over the standard direct encoding.
The conclusion is that, across evolutionary computation areas as diverse as genetic programming, neuroevolution, genetic algorithms, and theory, expressive encodings can be a key to understanding and realizing the full power of evolution.
\end{abstract}

\ccsdesc[500]{Computing methodologies~Genetic algorithms}
\ccsdesc[300]{Computing methodologies~Genetic programming}
\ccsdesc[300]{Computing methodologies~Neural networks}

\keywords{genetic algorithms, universal approximation, expressive encodings}


\maketitle

\section{Introduction}
\label{sec:introduction}

Evolutionary algorithms (EAs) promise to generate powerful solutions in complex environments.
To fulfill this promise, not only must great solutions exist somewhere in their search space, but EAs must be able to find them with non-vanishing probability even in high-dimensional, dynamic, and deceptive domains.
One way to deliver on this promise is to design more and more specialized operators to capture the structure of a particular domain.
Although this approach has led to many practical successes \cite{chicano2017optimizing, deb2016breaking, goldberg1994genetic, whitley2019next}, it is inherently limited by the human effort required to design such operators for each domain: That is, humans remain a bottleneck.

This paper advocates for an alternative approach: using simple, generic evolutionary operators with general genotype encodings in which arbitrarily high levels of complexity can accumulate.
Such encodings are termed \emph{expressive encodings} in this paper, and are analyzed theoretically and experimentally.
Using standard simple genetic operators (SGOs) involving only one or two parents, expressive encodings are capable of universal approximation of child phenotype distributions.  Thus, in theory, no hand-design is necessary to capture complex domain-adapted behavior.

Expressive encodings can be implemented with universal function approximators such as genetic programming and neural networks; the paper shows
that such systems can achieve up to super-exponential improvements in settings that are intractable with standard direct encodings.
The space of expressive encodings overlaps with indirect encodings, but is distinctly different: Some direct encodings can be expressive as well.

The expressive encoding approach can be seen as analogous to recent progress in machine learning:
Prior to deep learning \cite{goodfellow2016deep, lecun2015deep}, progress was made by crafting better features and increasingly complex learning algorithms to solve different kinds of problems.
With deep learning, large, complex structures with a general form (deep networks) allowed better solutions to be found across huge swaths of problems using a generic and simple algorithm: stochastic gradient descent (SGD).
EA could make similar progress via complex, general encodings evolved with SGOs.

The approach creates further opportunities in at least four areas of evolutionary computation:
\begin{itemize}
    \item Genetic Programming and Neuroevolution: Expressive encodings in these frameworks can be developed further and used as a starting point for improved methods.
    \item Genetic Algorithms: They can be defined for GAs in general, increasing their power.
    \item Theory: They can serve as a foundation for a new theoretical perspective on EAs.
\end{itemize}
In other words, not only are expressive encodings powerful and valuable in their own right, they can provide a shared platform of research that would allow fruitful communication and knowledge-sharing between often disparate subfields.

\section{Conceptual Framework}
\label{sec:preliminaries}

This section provides background on encodings and genetic operators, including running examples that will be used in the paper.

\subsection{Encodings}
\label{subsec:encodings}

An \emph{encoding} is a function $E: X \to Y$, where $X$ is a set of \emph{genotypes} and $Y$ is a set of \emph{phenotypes}.
The \emph{genotype} (or \emph{genome}) $x \in X$ is a description of the individual, which $E$ uses to produce the \emph{phenotype} $y \in Y$, which can then be evaluated in a given environment.
A fitness function $f: Y \to \mathbb{R}$ assigns a score to the phenotype.
The evolutionary computation literature is filled with a diversity of encodings.
This paper focuses on some of the most popular, and, for simplicity, primarily uses a boolean vector phenotype space $Y = \{0, 1\}^n$, but the results can be extended to other settings.

\subsubsection{Direct Encoding}
This is the simplest and most popular approach in using GAs and EAs for optimization.
With a direct encoding, $E$ is the identity function $E(x) = y$, and $X = Y$.
It is called \emph{direct} because the algorithm directly evolves elements of the phenotype: the genotype contains no information or structure not in the phenotype, and visa versa.
When evolving boolean vectors with a direct encoding, the genotype and phenotype are simply a vector of bits, which is the most common setting in EA theory \cite{auger2011theory, doerr2015black, doerr2012multiplicative, watson2007building, witt2013tight}.

\subsubsection{Neural Networks}
Evolving neural networks (NNs), or \emph{neuroevolution} (NE), is a popular approach to discovering solutions for problems like function approximation, control, and sequential decision-making \cite{fogel1990evolving, miikkulainen2019evolving, stanley2002evolving, stanley2019designing, such2017deep, yao1999evolving}.
The phenotype is usually a function, which may receive multiple distinct inputs during its evaluation.
Many encodings exist for evolving NNs. The simplest of them is a direct encoding, where the NN structure is fixed and its weights are evolved \cite{fogel1990evolving, such2017deep}.
However, NNs can also be used as the genetic encoding in cases where the phenotype is not a function but simply a fixed structure, such as a binary vector.
In this case, since there is no varying input to the phenotype during evaluation, a fixed value, e.g., \textbf{1}, can simply be fed into the network to generate the fixed phenotype.
Formally, an NN genotype $h$ is of the form $h: \mathbb{R}^1 \to \mathbb{R}^N$ and a sigmoid activation $\sigma$ in the final layer squashes the output of the NN into $(0, 1)$. Therefore, the overall encoding is
\begin{equation}
\label{eq:nn_encoding}
  E(h) = \textrm{round}(\sigma(h(\mathbf{1}))),  
\end{equation}
where `round' is applied elementwise, so that $E(h)$ produces a binary vector.
In this paper, all internal nodes have sigmoid activation, and all biases are 0 unless otherwise noted.

Although we are not aware of any prior work that has evolved NNs to optimize binary vectors, there is a substantial prior work evolving NNs to generate static artifacts such as pictures \cite{krolikowski2020quantum, liapis2021transforming, secretan2008picbreeder} and 3D objects \cite{clune2011evolving, lehman2016creative}.
Similarly to this paper, the motivation is that the structure of NNs can lead to interesting patterns in how offspring phenotypes relate to parent phenotypes.
High-level (and often interpretable \cite{huizinga2018emergence}) reproductive complexity can be achieved that would not be possible if, say, pixels or voxels were evolved directly.
The complexity of what is possible in evolution at any point in time is accumulated in what has been evolved so far.
This area of work has achieved impressive, visually appealing results \cite{lehman2012beyond, nguyen2015innovation}.
A goal of the present work is to show that such an approach is indeed fundamentally powerful for problem-solving, and overcomes serious fundamental limitations of direct encodings.


\subsubsection{Genetic Programming}
Like neuroevolution, genetic programming (GP) is used in situations where phenotypes require complex behavior \cite{brameier2007linear, hodjat2014maintenance, koza1992genetic, langdon2013foundations, o2013ec, spector2002genetic, shahrzad2015tackling}.
GP is often used for function approximation \cite{orzechowski2018we, schmidt2009distilling}, but its main motivation is to generate programs that meet a given description.
As evolved programs accumulate complexity, like in the case of NNs \cite{huizinga2018emergence,lehman2013evolvability}, the behavior of evolution itself changes over time \cite{altenberg1994evolution,banzhaf2014genetic,hu2011robustness}.
So, like NN, GP can be used to generate static structures, leading to benefits from the complexity of evolved programs.
As there are many languages for human programmers, there are many encodings for GP \cite{brameier2007linear, koza1992genetic, spector2002genetic}.
The encoding can generally be defined by the available terminals and operators.
This paper considers a small set of terminals and operators, which can naturally be extended:
\begin{itemize}
    \item Terminals: binary vectors with evolved values, integers;
    \item Operators: $<$, $>$, $+$, $\parity$, $\oplus$, if, return.
\end{itemize}
The binary vectors found at terminals can be of varying length, but at least one must be of length $n$ so that the program can return a solution of length $n$.
Vectors of length 1 will be broadcast when used in binary operations;
`$<$' and `$>$' compare vectors from left to right as if they were binary integers, e.g., $([0, 1, 0] < [0, 1, 1]) = (010 < 011) = 1$;
`par' returns the parity of the vector; 
`$\oplus$' performs elementwise addition (mod 2).
A program using these operators can be rendered in sequential, tree, or graph form \cite{brameier2007linear,koza1992genetic}; this paper uses sequential programs for readability.\\


\noindent Of course, NNs and programs on their own are expressive in the space of functions: NNs can approximate any function arbitrarily closely \cite{cybenko1989approximation, hornik1989multilayer, kolmogorov1957representation}, and sufficiently powerful programming languages can compute anything that is computable \cite{church1936unsolvable, spector2002genetic, turing1936computable}.
This paper focuses on a different type of expressivity: How do different encodings enable the evolutionary process to behave in complex and powerful ways?
This behavior involves pairing encodings with genetic operators, which are discussed next.

\subsection{Genetic Operators}
\label{subsec:genetic_operators}

Genetic (or evolutionary) operators are the mechanisms by which genomes reproduce to generate other genomes.
A genetic operator $g$ is a (usually stochastic) function that produces a new genome $x'$ given a set of $n_g$ parent genomes $X_p \subset X$.
Since $g(x)$ results in a distribution over genomes, we can write $x' \sim g(X_p)$.
This paper focuses on the two most common operators (in both nature and computation): crossover and mutation.
For consistency, assume the genotype can be flattened into a string of symbols in a canonical way, so that $x^j$ refers to the $j$th symbol in the string form of a genotype $x$.
The following operators are likely familiar to EA practitioners, but they are briefly described here for completeness.

\subsubsection{Uniform Crossover}
This operator $g_c$ takes two parents of equivalent structure and produces a child by independently selecting the value in one of the parents at random for each element.
That is, $x_c \sim g_c(x_1, x_2) \implies x_c^j \sim U(\{x_1^j, x_2^j\}) \ \forall \ j$, where $U$ is the uniform distribution.
Importantly, if the two parents have the same value at index $j$, then the child is also guaranteed to have that value at $j$: $x_1^j = x_2^j \implies x_1^j = x_c^j$.
The results on uniform crossover in this paper should be extendable to other forms of crossover, including single- and multi-point crossover \cite{de1992formal, mitchell1991royal, watson2007building}.


\subsubsection{Single-point Mutation}
With single-point mutation $g_m$ the child is a copy of a single parent with a single location altered.
For example, if the encoding is an NN, $g_m$ can alter a single weight in the network, e.g., by adding Gaussian noise.
If the encoding is GP, $g_m$ can replace a symbol with a different valid symbol, e.g., flip a bit of a location that contains a binary value.
Single-point mutation is similar to uniform mutation, which mutates each element independently with equal probability, but is simpler to analyze, and has been shown to be effective in many benchmarks \cite{doerr2008comparing}.

\subsubsection{Simple Genetic Operators (SGOs)}
We call a genetic operator \emph{simple} if both its description length and the cardinality of its parent set are constant (w.r.t. the phenotype dimensionality $n$) .
Clearly, the above examples are simple, which matches our intuition, since they are some of the most basic operators in EAs.
These operators are also simple in the colloquial sense: They are easy to explain, implement, and apply in a wide variety of settings without invoking too much domain-specific or encoding-specific complexity.
Operators that are not simple include model-based EAs such as EDAs \cite{harik1999compact,hauschild2011introduction,pelikan1999boa}, linkage-based algorithms \cite{goldman2014parameter,thierens2010linkage}, and evolutionary strategies \cite{hansen2006cma,wierstra2014natural}, which construct $\Omega(n)$-size probabilistic models from $\Omega(n)$ genotypes, and use these models to generate new candidates.
Such models usually have a restricted structure, but in theory could model any phenotype distribution.
This paper shows that SGOs are also fundamentally powerful, as long as the encoding is sufficiently complex.
SGOs are more in line with how evolution operates in nature, as well as the original motivation for GAs \cite{holland1992genetic}. They also avoid the pitfalls of a central algorithmic bottleneck:
The operations of variation in the algorithm can be exceedingly simple, and can still result in powerful complex behavior via complexity accumulating in genotypes.
This idea is made formal in the next section.

\begin{figure*}
    \fontsize{7.6}{7.6}\selectfont
    {\centering
    \begin{minipage}{0.21\linewidth}
        \begin{tcolorbox}[boxsep=5pt, left=8pt, right=5pt, top=1pt, bottom=1pt]
            \fontsize{8}{8.8}\selectfont
        \begin{verbatim}
a = 0
b = 1
if a + b > 1:
    return [1,...,1]
else:
    return [0,...,0]\end{verbatim}
        \end{tcolorbox}
        \centering{Parent 1}
    \end{minipage}
        \hspace{5pt}
    \begin{minipage}{0.21\linewidth}
        \begin{tcolorbox}[boxsep=5pt, left=8pt, right=5pt, top=1pt, bottom=1pt]
            \fontsize{8}{8.8}\selectfont
        \begin{verbatim}
a = 1
b = 0
if a + b > 1:
    return [1,...,1]
else:
    return [0,...,0]\end{verbatim}
        \end{tcolorbox}
        \centering{Parent 2}
    \end{minipage}
    \hspace{5pt}
    \begin{minipage}{0.35\linewidth}
    \centering
    \includegraphics[width=0.675\linewidth]{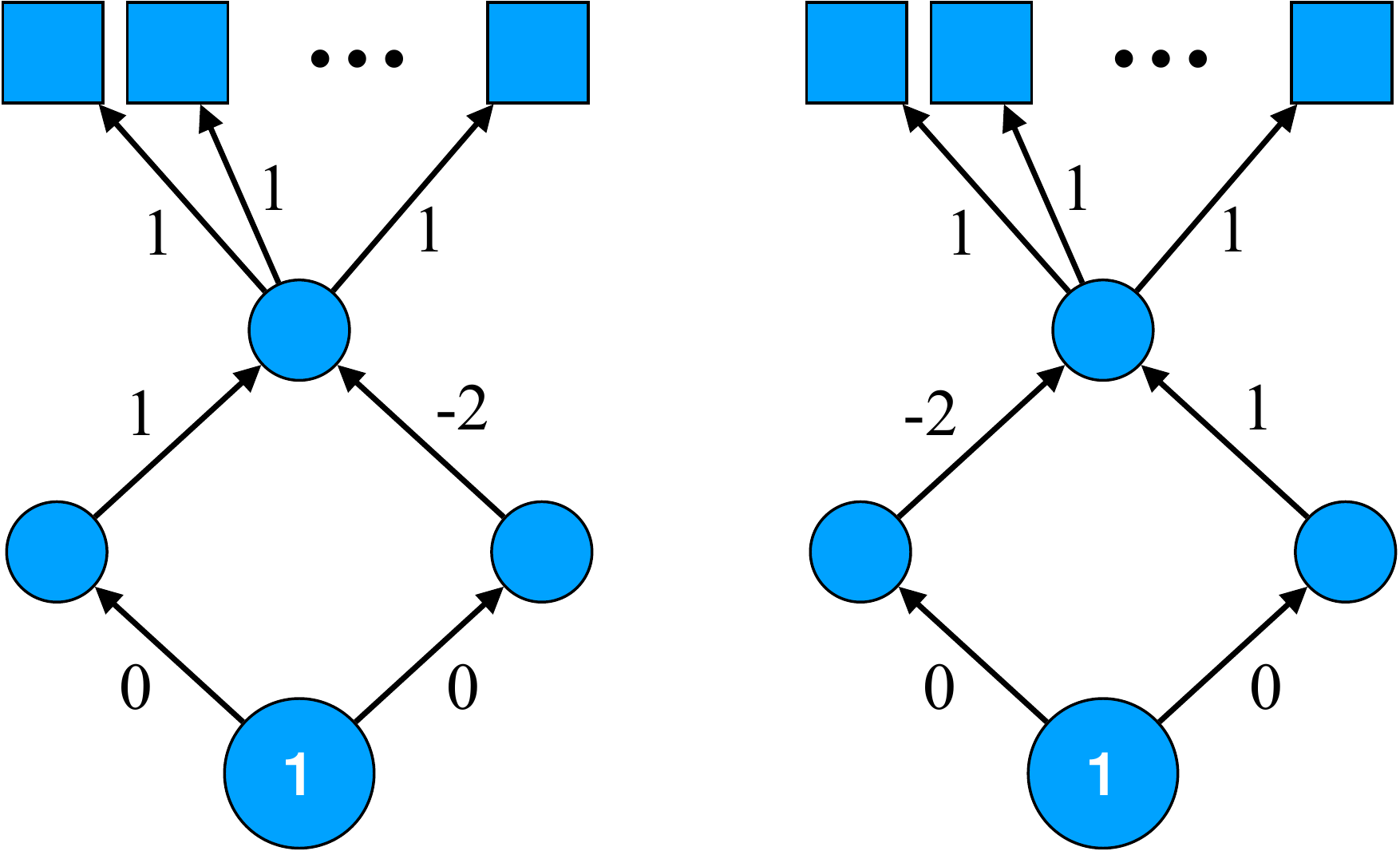}\\
    Parent 1 \hspace{38pt} Parent 2
    \end{minipage}
    \begin{minipage}{0.15\linewidth}
    \centering
    Parent 1 \vspace{8pt}\\
    \includegraphics[width=0.8\linewidth]{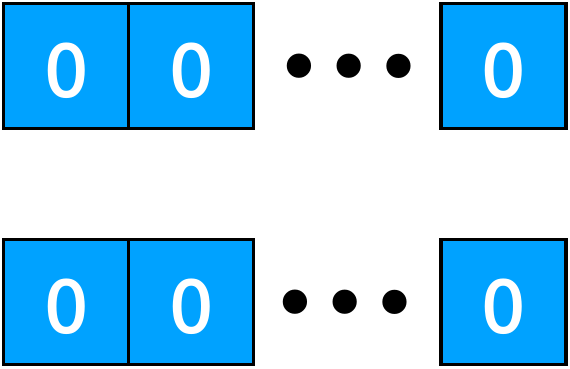}\\
    \vspace{5pt}Parent 2
    \end{minipage}}\\
    \raggedright\hspace{117pt}(a)\hspace{203pt}(b)\hspace{121pt}(c) \vspace{-5pt}
    \caption{\emph{Miracle Jump Parents.} 
    (a) Two GP parents whose phenotypes are all 0's, but whose crossover has maximal jump (to a child of all 1's) with probability 0.25.
    They differ only in their values of $a$ and $b$; the probability is independent of phenotype dimensionality.
    (b) Two NN parents with this same property;
    They differ only in the weights in the second layer.
    (c) Directly encoded parents cannot have this property:
    If both parents are all 0's, their crossover cannot yield all 1's.
    This minimal example illustrates how expressive encodings yield high-dimensional structured behavior that direct encodings cannot capture. 
    }
    \label{fig:miracle_jump}
\end{figure*}

\section{Expressive Encodings}
\label{sec:expressive_encodings}

The behavior of evolutionary algorithms can be described by their \emph{transmission function}, which probabilistically maps parent phenotypes to child phenotypes \cite{cavalli1976evolution,slatkin1970selection,altenberg1994evolution}.
EAs that use expressive encodings can, in principle, yield the behavior of \emph{any} transmission function, making them general and powerful generative systems.
\begin{definition}[Expressive Encoding]
An encoding $E: X \to Y$ is \emph{expressive} for a simple genetic operator $g$ if, for any set of parent phenotypes $\{y_1,\ldots,y_{n_g}\} = Y_p \subset Y$, any probability density $\mu$ over $Y$, and any $\epsilon > 0$, there exists a set of parent genotypes $\{x_1,\ldots,x_{n_g}\} = X_p \subset X$ such that $E(x_i) = y_i \ \forall \ y_i \in Y_p$, and
\begin{equation}
    \Big| \ \Pr \big[E(g(X_p)) = y\big] - \mu(y) \ \Big| < \epsilon \ \ \forall y \in Y.
\end{equation}
\end{definition}
In other words, starting from any initial set of parent phenotypes, a single application of the genetic operator can generate any distribution of child phenotypes.
This paper focuses on the case where $\mu$ is a discrete distribution, but the ideas can be extended to the continuous case.
The \emph{complexity} of an expressive encoding with a particular genetic operator is the genome size required to achieve the desired level of approximation.
Expressive encodings have the satisfying property that the current definition of the evolutionary process must be stored in the genomes themselves.
There is no reliance on a large external controller of evolution; the only external algorithmic information is in SGOs; the process and artifacts of evolution are stored at a single level: in the population.

It may seem that all expressive encodings are \emph{indirect encodings} \cite{stanley2003taxonomy,stanley2009hypercube,stanley2019designing}.
However, if the phenotype space is sufficiently expressive, such as in the case of searching for a neural network to perform a task, then \emph{a direct encoding may be an expressive encoding}.
This property is demonstrated for the case of neural networks in the next section (see Theorem~\ref{thm:nn_direct}).
There are other kinds of systems capable of universal approximation of probability distributions, such as neural generative models (e.g., the generator in a GAN architecture \cite{goodfellow2014generative,lu2020universal}).
If EAs were not capable of such universality, then one should be hesitant to use them as a problem-solving tool or engine of a generative system.
Fortunately, they are, as is shown next.

\section{Expressivity of GP, NN, Direct, and Universal Approximator Encodings}
\label{sec:sampling_and_universality}

This section provides constructions that illustrate the fundamental power of expressive encodings, including showing that NNs and GP are expressive for uniform crossover and one-point mutation, while the direct encoding is not.
A key mechanism is representation of switches in the genome. This mechanism is first introduced in the special case of analyzing big jumps in GP, NN, and direct encoding, and the results are then generalized to full universal approximation.

\subsection{Special Case: Miracle Jumps}

There is a common hope in evolutionary computation of stumbling upon `big jumps' that end up being useful.
There is a large theory literature on how well algorithms handle `jump functions', i.e., problems where there are local optima from which a jump must be taken to reach the global optimum \cite{bambury2021generalized,dang2016escaping,whitley2018exploration}.
This subsection shows how expressive encodings can enable reliable jumps of maximal size---a useful feature for EAs in general.
We call this behavior a \emph{miracle jump} because, if it were to occur in a standard evolutionary algorithm with a direct encoding, it would be considered a miracle.

Let $Y = \{0, 1\}^n$.
Given an encoding $E$ and genetic operator $g$, the goal is to find $X_p$ such that $E(x) = [0, \ldots, 0] \ \forall x \ \in X_p$ and $\Pr [ g(X_p) = [1, \ldots, 1]] = \Theta(1).$
In other words: Find parents whose phenotypes are all zeros, but who generate a child of all ones with probability that does not go to zero as $n$ increases.
To solve this problem, the encoding must be capable of encoding a kind of switch that allows non-trivial likelihood of flipping to the all ones state.

\subsubsection{Genetic Programming}

Consider the two parent programs shown in Figure~\ref{fig:miracle_jump}(a).
The two parents are equivalent except for their values of $a$ and $b$.
So, uniform crossover $g(X_p)$ results in a $\nicefrac{1}{4}$ chance that $a = b = 1 \implies E(g(X_p)) = [1, \ldots, 1]$.

\subsubsection{Neural Networks}

Consider the two parent NNs shown in Figure~\ref{fig:miracle_jump}(b).
The two parents are equivalent except for the values of each of their two weights in the first layer.
With uniform crossover, there is a $\nicefrac{1}{4}$ chance the weights of the first layer of the child are all $1$, resulting in phenotype of $[1, \ldots, 1]$.

\subsubsection{Direct Encoding}

Since there are no 1's in either parent (Figure~\ref{fig:miracle_jump}(c)), this solution is impossible for crossover to find. Single-point mutation only changes a single bit, but even uniform mutation fails: It flips bits i.i.d. with probability $p < 1$, and $\lim_{n \to \infty} p^n = 0.$\\

\noindent The GP and NN constructions above are built on the idea of having a large part of the genome being completely shared between the two parents, and then having some auxiliary values that are unshared and function as control bits that have a high-level influence on how the phenotype is generated.
This kind of construction is used to generalize these results in the next subsection.

\subsection{General Case: Universal Approximation}

This subsection shows that certain encodings are expressive with SGOs.
Unless otherwise noted, $Y$ is the phenotype space of binary vectors $Y = \{0, 1\}^n$.
Let $\by_i \ (\forall \ i \in [1, \ldots, m])$ be the desired phenotypes in the child distribution, and $p_i$ be their associated probabilities.
Notice that, since there can be no more than $\frac{1}{\epsilon}$ $p_i$'s such that $p_i \geq \epsilon$, any construction need only assign nonzero probability to at most the top $\frac{1}{\epsilon}$ most probable $\by_i$'s to achieve an approximation error of $\epsilon$. 
While expressivity can be demonstrated with many different constructions, the goal of this section is to provide constructions that are both intuitive and highlight where the power of expressive encodings comes from.  Sketches of the proofs are provided below. Detailed proofs are included in Appendix~\ref{sec:proofs_for_universality}.

\begin{figure*}
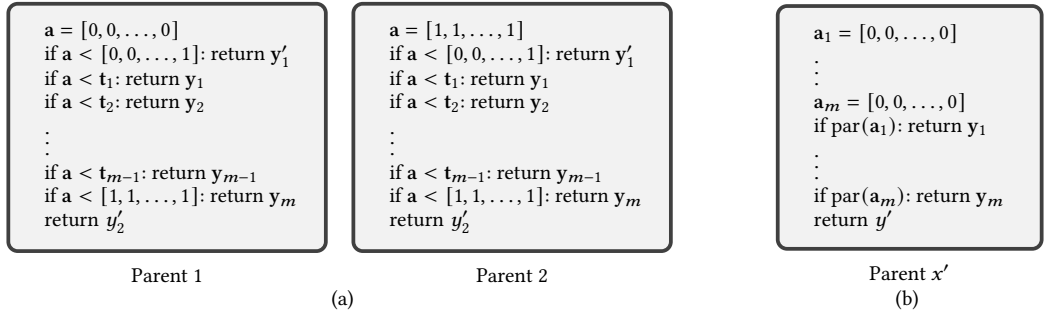

    \centering
    \fontsize{8}{9}\selectfont
    \begin{minipage}{0.24\linewidth}
    \begin{tcolorbox}[boxsep=5pt, left=8pt, right=5pt, top=1pt, bottom=1pt]
$\ba = [0, 0, \ldots, 0]$\\
if $\ba < [0, 0, \ldots, 1]$: return $\by'_1$\\
if $\ba < \bt_1$: return $\by_1$\\
if $\ba < \bt_2$: return $\by_2$\\
$\vdots$\\
if $\ba < \bt_{m-1}$: return $\by_{m-1}$\\
if $\ba < [1, 1, \ldots, 1]$: return $\by_m$\\
return $y'_2$
    \end{tcolorbox}
    \centering{Parent 1}
    \end{minipage}
    \hspace{5pt}
    \begin{minipage}{0.24\linewidth}
    \begin{tcolorbox}[boxsep=5pt, left=8pt, right=5pt, top=1pt, bottom=1pt]
$\ba = [1, 1, \ldots, 1]$\\
if $\ba < [0, 0, \ldots, 1]$: return $\by'_1$\\
if $\ba < \bt_1$: return $\by_1$\\
if $\ba < \bt_2$: return $\by_2$\\
$\vdots$\\
if $\ba < \bt_{m-1}$: return $\by_{m-1}$\\
if $\ba < [1, 1, \ldots, 1]$: return $\by_m$\\
return $y'_2$
    \end{tcolorbox}
    \centering{Parent 2}
    \end{minipage}
    \hspace{35pt}
    \begin{minipage}{0.20\linewidth}
    \begin{tcolorbox}[boxsep=5pt, left=8pt, right=5pt, top=1pt, bottom=1pt]
$\ba_1 = [0, 0, \ldots, 0]$\\
$\vdots$\\
$\ba_m = [0, 0, \ldots, 0]$\\
if $\parity(\ba_1)$: return $\by_1$\\
$\vdots$\\
if $\parity(\ba_m)$: return $\by_m$\\
return $y'$
    \end{tcolorbox}
    \centering{Parent $x'$}
    \end{minipage}\\ \vspace{2pt}
    \raggedright\hspace{180pt}(a)\hspace{205pt}(b)
    \caption{\emph{Universal GP Parents.}
    (a) A template for two parents whose crossover can approximate any probability distribution over phenotypes.
    The parents differ only in their values of the variable $\ba$;
    (b) A similar template for which single-point mutation approximates any probability distribution over phenotypes.
    Each $\ba_i$ may have a different length, and `par' indicates the parity function.
    These templates can be used to show that GP is an expressive encoding.
    }
    \label{fig:gp_universal}
\end{figure*}

\begin{thm}
Genetic programming is an expressive encoding for uniform crossover, with complexity $O(mn - m\log \epsilon)$. \label{thm:gp_universal_crossover}
\end{thm}
\begin{pfs}
Take the parents in Figure~\ref{fig:gp_universal}(a).
They differ only in their values of $\ba$, and their child's value can be viewed as an integer sampled uniformly from $\{0, 2^{\dim(\ba)} - 1\}$.
If $\dim(\ba) > \lceil \lg \frac{1}{\epsilon} + 1 \rceil$, then the $\bt_i$'s can be chosen so that $\mu(\by_i)$ probability is apportioned to each $\by_i$ with error less than $\epsilon$. \qed
\end{pfs}


\begin{thm}
\label{thm:gp_mutation}
Genetic programming is an expressive encoding for single-point mutation, with complexity $O(\frac{mn}{\epsilon})$.
\end{thm}
\begin{pfs}
Take the parent in Figure~\ref{fig:gp_universal}(b).
Let $\dim(\ba_i)$ be proportional to $\mu(\by_i)$, and then scale up all $\dim(\ba_i)$ so the chance of a mutation in a $\by_i$ is sufficiently small.
\qed \vspace{7pt}
\end{pfs}

Notice that these constructions resemble the Pitts GP model \cite{hodjat2014maintenance,o2013ec,smith1980learning,urbanowicz2009learning}.
Neither of the above constructions depend on the $\by_i$'s being drawn from a particular phenotype space.
The $\by$'s could have any structure and be drawn from any phenotype space, and a similar construction could be provided, highlighting the fact that expressive encodings enable powerful EA behavior that can be general across problem domains.

\begin{figure*}
    \centering
    \begin{minipage}{0.5\linewidth}
    \includegraphics[width=\linewidth]{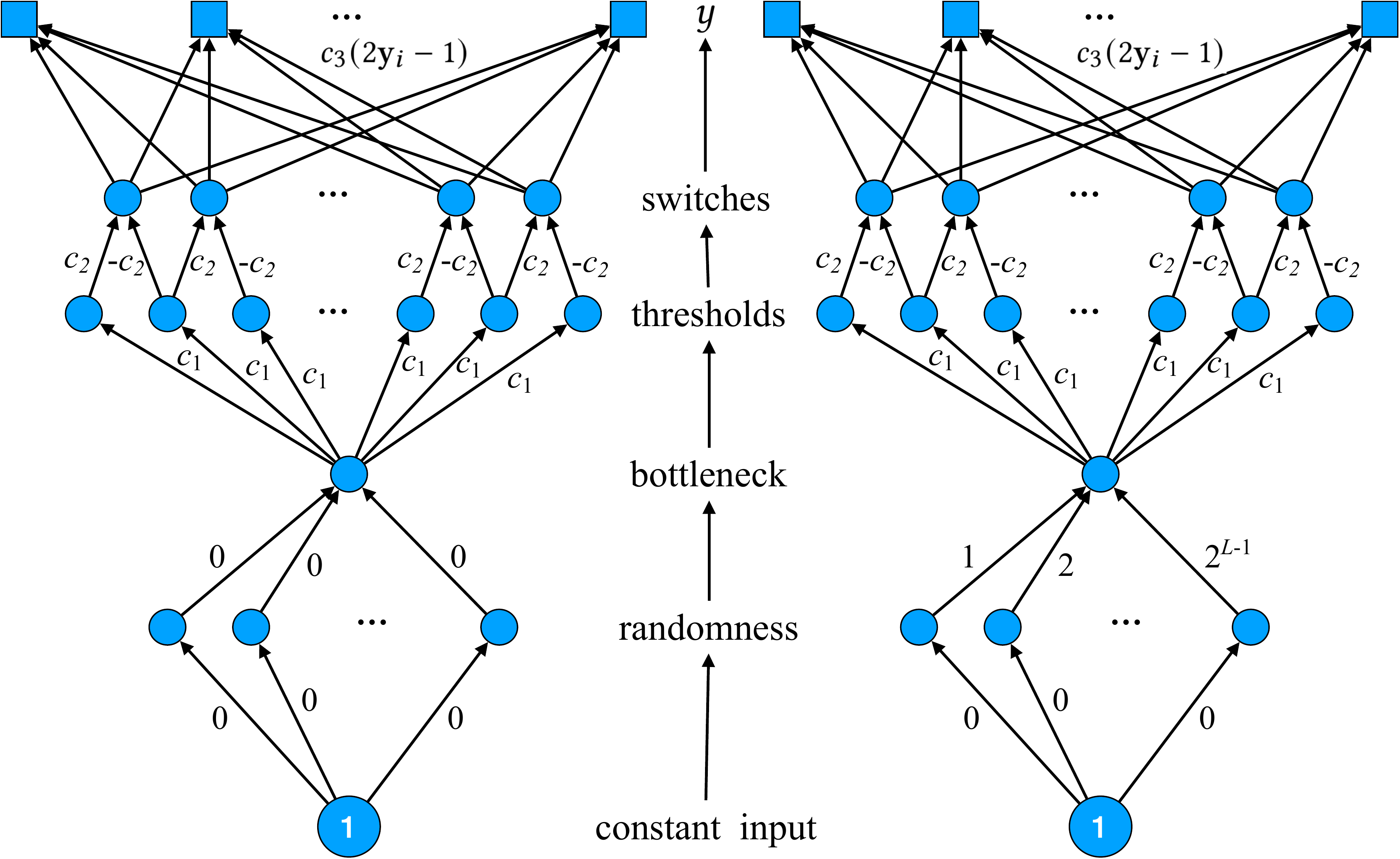}\\ \vspace{5pt}
    \raggedright\hspace{47pt}Parent 1 \hspace{98pt} Parent 2
    \end{minipage}
    \hspace{20pt}
    \begin{minipage}{0.45\linewidth}
    \includegraphics[width=\linewidth]{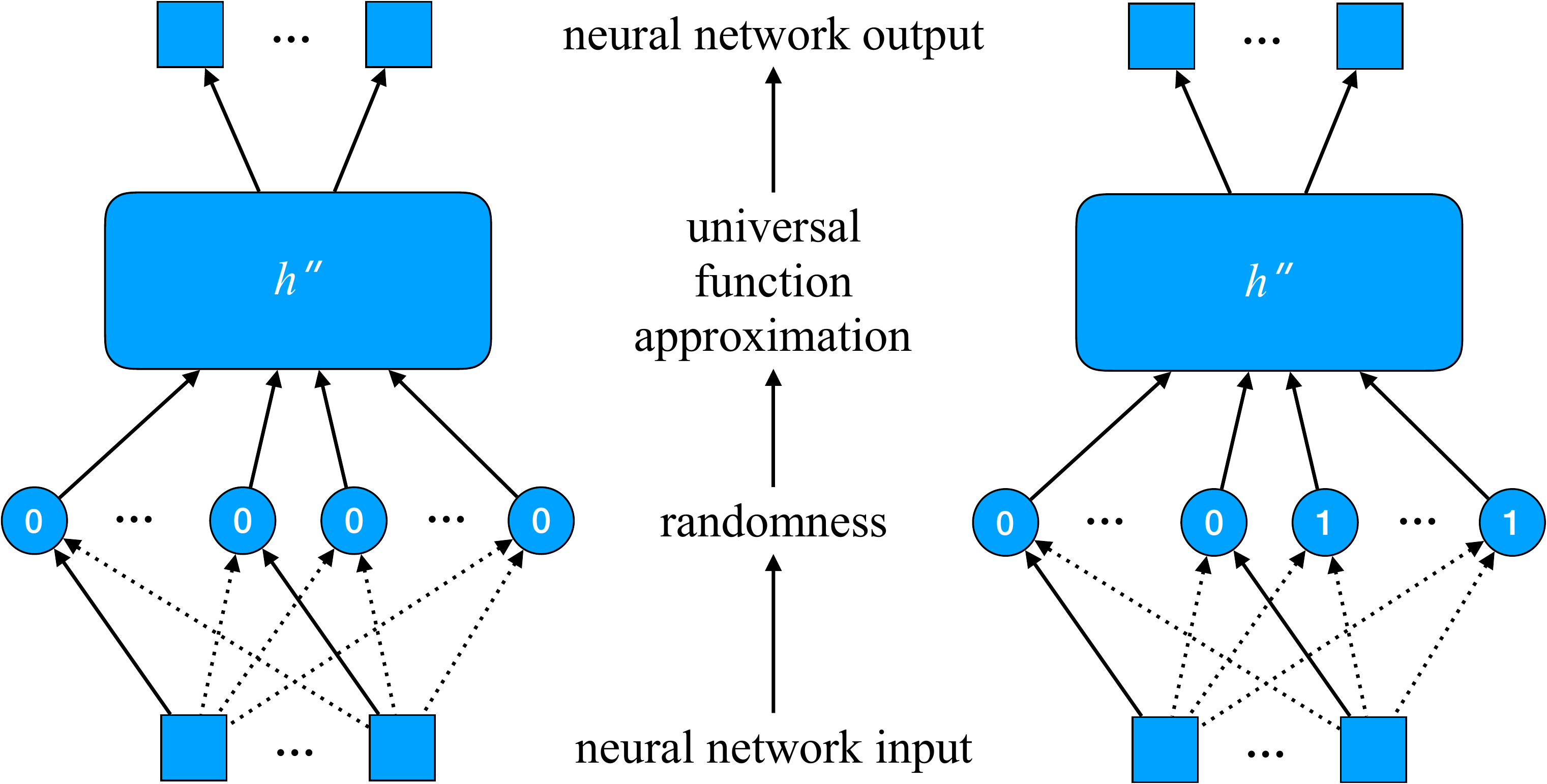}\\ \vspace{8pt}
    \raggedright\hspace{27pt}Parent 1 \hspace{110pt} Parent 2
    \end{minipage}\\
    \raggedright\hspace{124pt}(a)\hspace{256pt}(b)
    \caption{\emph{Universal NN Parents.}
    (a) A class of parent NNs that can approximate a given probability distribution over phenotypes when recombined with uniform crossover (Theorem~\ref{thm:nn_crossover}). Values are chosen for $c_1$, $c_2$, and $c_3$ such that for the desired phenotypes there are mutually-exclusive switches that fire with the corresponding probabilities.
    Parent 2 can also be used as the parent in the case of mutation (Theorem~\ref{thm:nn_mutation}).
    (b) A direct encoding of two NN parents that can approximate any probability distribution over NN functions when recombined with uniform crossover.
    Solid lines indicate a weight of 1, dotted lines 0.
    The numbers in the units of the first hidden layer indicate the bias of that unit.
    A generated child will have biases sampled uniformly from $\{0, 1\}^L$, and the remainder of the network, $h''$, decides which function the entire network computes based on the sampled values.
    As $L$ increases, $h''$ can better approximate the desired distribution.
    Thus, NNs can implement an expressive encoding, even when they themselves are based on a direct encoding.
    }
    \label{fig:nn_universal}
\end{figure*}

Note also that the construction for crossover is asymptotically more compact than that for mutation: $-m\log \epsilon = m\log \frac{1}{\epsilon}$ vs. $\frac{m}{\epsilon}$ in the first term. This result suggests that crossover possibly has an inherent advantage over mutation: It achieves equivalently complex behavior with a more compact genome.
An interesting question is: Is this observation reflected in biological evolution?
Importantly, the same phenomenon emerges for NNs:

\begin{thm}
\label{thm:nn_crossover}
Feed-forward neural networks with sigmoid activation are an expressive encoding for uniform crossover, with complexity $O(mn - \log \epsilon)$.
\end{thm}
\begin{pfs}
Take the parents in Figure~\ref{fig:nn_universal}(a), who differ only in their second-layer weights.
The input to the child's bottleneck is proportional to a uniformly-sampled integer.
Choosing threshold biases and high enough $c_1$, $c_2$, and $c_3$ results in a mutually-exclusive switch for each $\by_i$ that fires with the correct probability. \qed \vspace{7pt}
\end{pfs}

In the case of NNs, let single-point mutation be a Gaussian mutation, i.e., a single weight is selected uniformly at random and modified by the addition of Gaussian noise.
The expressivity construction is similar to that of crossover, leading to the following:

\begin{thm}
\label{thm:nn_mutation}
Feed-forward neural networks with sigmoid activation are an expressive encoding for single-point mutation, with complexity $O(\frac{mn}{\epsilon})$.
\end{thm}
\begin{pfs}
Take Parent 2 from Figure~\ref{fig:nn_universal}(a).
Choose $L$ so that mutation is very likely to occur in the first two layers. Since applying $g_m$ there yields \emph{some} continuous distribution over bottleneck output, suitable thresholds and switches can be created. \qed \vspace{7pt}
\end{pfs}

The constructions so far give concrete ways to create parents to demonstrate expressivity.
The next result generalizes these ideas to encodings built from any universal function approximator.
The idea is that any universal function approximator can be extended to an expressive encoding via $E_\Omega$, defined as follows:

\begin{definition}[$E_\Omega$]
\label{def:ufa}
Let $\Omega$ be any universal function approximator.
Define $E_\Omega$ to be an encoding whose genotypes are of the form $\omega(\va)$, where $\va \in \{0, 1\}^L$, and $\omega \in \Omega$ is a function $\omega: \{0, 1\}^L \to Y$.
\end{definition}

\begin{thm}
\label{thm:ufa}
$E_\Omega$ is an expressive encoding for uniform crossover.
\end{thm}
\begin{pfs}
Choose a large enough $L = \Theta(-\lg \epsilon)$.
Then, let $x'_1 = \omega([0,\ldots,0])$ and $x'_2 = \omega([1,\ldots,1])$, for a suitable $\omega \in \Omega$.
\qed \vspace{7pt}
\end{pfs}
So, if you have a favorite class of models that is expressive in terms of its function approximation capacity, it can be turned into a potentially powerful evolutionary substrate. 


The constructions so far have used \emph{indirect encodings}, in that the phenotype space (e.g., binary vectors) is of a different form than the genotype space (e.g., GP, NN, or $\omega(\va)$).
However, an encoding need not be indirect to be expressive, as is demonstrated by the following case of direct encoding of neural networks.

\begin{thm}
\label{thm:nn_direct}
Direct encoding of feed-forward neural networks with sigmoid activation is an expressive encoding for uniform crossover.
\end{thm}
\begin{pfs}
Take the parents in Figure~\ref{fig:nn_universal}(b).
They differ only in the biases in the first layer of nodes.
If this layer is large enough, $h''$ has enough information from a child's biases to decide which function the overall NN should compute.
\qed \vspace{7pt}
\end{pfs}

This section has shown that sufficiently complex evolutionary encodings, in particular NN and GP, are expressive.
Another way of seeing the power of this property is to consider any stochastic process (e.g., EA) that samples from distribution $\mu$ after $T$ steps: An expressive encoding can simulate this behavior \emph{in one step}.
This view gives us an analogy to a powerful result from the NN meta-learning literature: Given a distribution of tasks, there exists a neural network that can learn any task from this distribution with a single step of gradient descent \cite{finn2017meta}.
This connection is meaningful: while meta-learning should be able to encode any learning process, evolution should be able to encode any phenotype sampling process.

The fact that the above encodings are expressive with single-point mutation, also known as \emph{random local search} \cite{doerr2008comparing}, is remarkable.
Thanks to expressive encoding, random \emph{local} search in the genotype space leads to maximally \emph{global} evolutionary sampling in the phenotype space.
The next section will show that this property can be used to solve challenging problems where the standard direct encoding is likely to fail.


\section{Adaptation and Convergence}
\label{sec:adaptation_and_convergence}

Section~\ref{sec:sampling_and_universality} showed how arbitrarily complex behavior is possible with a single application of an SGO when encodings are expressive.
An immediate question is: Can this power actually be exploited to solve challenging problems with evolution?
This section considers some of the most frustrating kinds of problems for the standard GA: those where there is high-level high-dimensional structure that can only be exploited if the population develops to a point where particular high-dimensional jumps emerge with reliable probability.
Standard GAs struggle to achieve such an emergence, since, as the dimensonality of the required jump increases, its probability vanishes.
This section considers three problems that cover the different ways in which such high-level high-dimensional structure can appear: It can be temporal and deterministic, temporal and stochastic, or spatial.
In all three cases, expressive encodings, implemented through GP, yield striking improvements over direct encodings.

\subsection{Problem Setup}

As is common for ease of analysis \cite{doerr2020probabilistic,lengler2020drift,meyerson2019modular,witt2013tight}, this section considers the (1+$\lambda$)-EA, with $\lambda \in \{1, 2\}$ (pseudocode is provided in Appendix~\ref{sec:algorithm1}).
In this algorithm, during each generation $\lambda$ candidates are independently generated by mutating the current champion. One of the best candidates replaces the champion if it meets the acceptance criteria, which is usually that it has higher fitness.
When the fitness function is dynamic, the fitness of the champion is evaluated in every generation along with the fitnesses of the candidates.

To evaluate algorithms on dynamic fitness functions, the concept of \emph{adaptation complexity} needs to be defined.
\begin{definition}[Adaptation Complexity]
The \emph{adaptation complexity} of an algorithm $A$ on a dynamic fitness function $f$ is the expected time $A$ spends away from a global optimum vs. at a global optimum in the limit, i.e., the proportion of time it spends \emph{adapting}.
Formally, if $f^*_t$ is the maximum value of $f$ at time $t$, and $f(E(x_t))$ is the best fitness in the population at time $t$, the adaptation complexity is
\begin{equation}
    \E\Bigg[\lim_{t \to \infty} \frac{\lvert \{t' < t : f(E(x_{t'})) \neq f^*_{t'}\} \rvert}{\lvert \{t' < t : f(E(x_{t'})) = f^*_{t'}\} \rvert}\Bigg] .
\end{equation}
\end{definition}

The three problems are considered in order of ease of analysis.
It may seem counter-intuitive to consider dynamic fitness functions before static fitness functions, but it turns out that the dynamism of the fitness function provides an exploratory power that expressive encodings are able to exploit naturally, but direct encodings cannot exploit at all (other advantages of dynamic fitness have been observed in prior work \cite{crombach2008evolution, kashtan2007varying}). As before, sketches of the proofs are provided in this section; details are included in Appendix~\ref{sec:proofs_for_convergence}.

\subsection{Challenge Problems}

\begin{problem}
\textbf{Deterministic Flipping Challenge (DFC):}
Take any two target phenotypes $\vy_1^*, \vy_2^* \in \{0, 1\}^n$, where $\vy_2^*$ is the complement of $\vy_1^*$.
At time $t=0$, the current target vector is $\vy_1^*$.
The fitness is the number of bits in the phenotype that match the target.
If the fitness of the champion is $n$, then at the next time step the current target vector flips to the other target vector. \vspace{5pt}
\end{problem}

The difficulty in this problem is clear: As soon as the maximum fitness is achieved, the target changes.
This kind of situation arises naturally in continual evolution, in which new problems arise over time, and the algorithm is initialized at its solution to the previous problem \cite{braylan2016reuse,fernando2017pathnet,francon2020effective,luders2016continual,neumann2015runtime,wang2020enhanced}.
For example, imagine a neural architecture search system in which new machine learning problems arise over time. The algorithm cannot be restarted from scratch for each problem, because it is so expensive to run, so it picks up from where it left off.
Partial attempts at such a system have been made previously \cite{golovin2017google}.
Problem 1 can also be viewed as a dynamic version of the OneMax problem;
other such versions have been considered in the past \cite{droste2003analysis,jansen2005theoretical,sudholt2018robustness}, but none allow for this extreme of a change in the target vector.
More generally, this test problem evaluates an algorithm's ability to avoid `catastrophic forgetting': it should not be too quick to forget useful past solutions, and ideally, it should be able to recover them in constant time.

\begin{figure*}
    \centering
    \fontsize{8}{9}\selectfont
\begin{minipage}{0.18\textwidth}
\centering
\begin{tcolorbox}[boxsep=5pt, left=8pt, right=1pt, top=1pt, bottom=1pt]
if $\parity(\mathbf{a})$: return $\mathbf{b}$\\
return $\mathbf{c}$
\end{tcolorbox}
(a)\\ \vspace{2pt}
\begin{tcolorbox}[boxsep=5pt, left=8pt, right=1pt, top=1pt, bottom=1pt]
$\mathbf{y} = \mathbf{0}$\\
if $\parity(\mathbf{a}_1)$: $\mathbf{y} \mathrel{\oplus}= \mathbf{b}$\\
if $\parity(\mathbf{a}_2)$: $\mathbf{y} \mathrel{\oplus}= \mathbf{c}$\\
return $\mathbf{y}$
\end{tcolorbox}
(b)
\end{minipage}
\hspace{5pt}
\begin{minipage}{0.26\textwidth}
    \centering
    (c) DFC\\ \vspace{-2pt}
    \includegraphics[width=\linewidth]{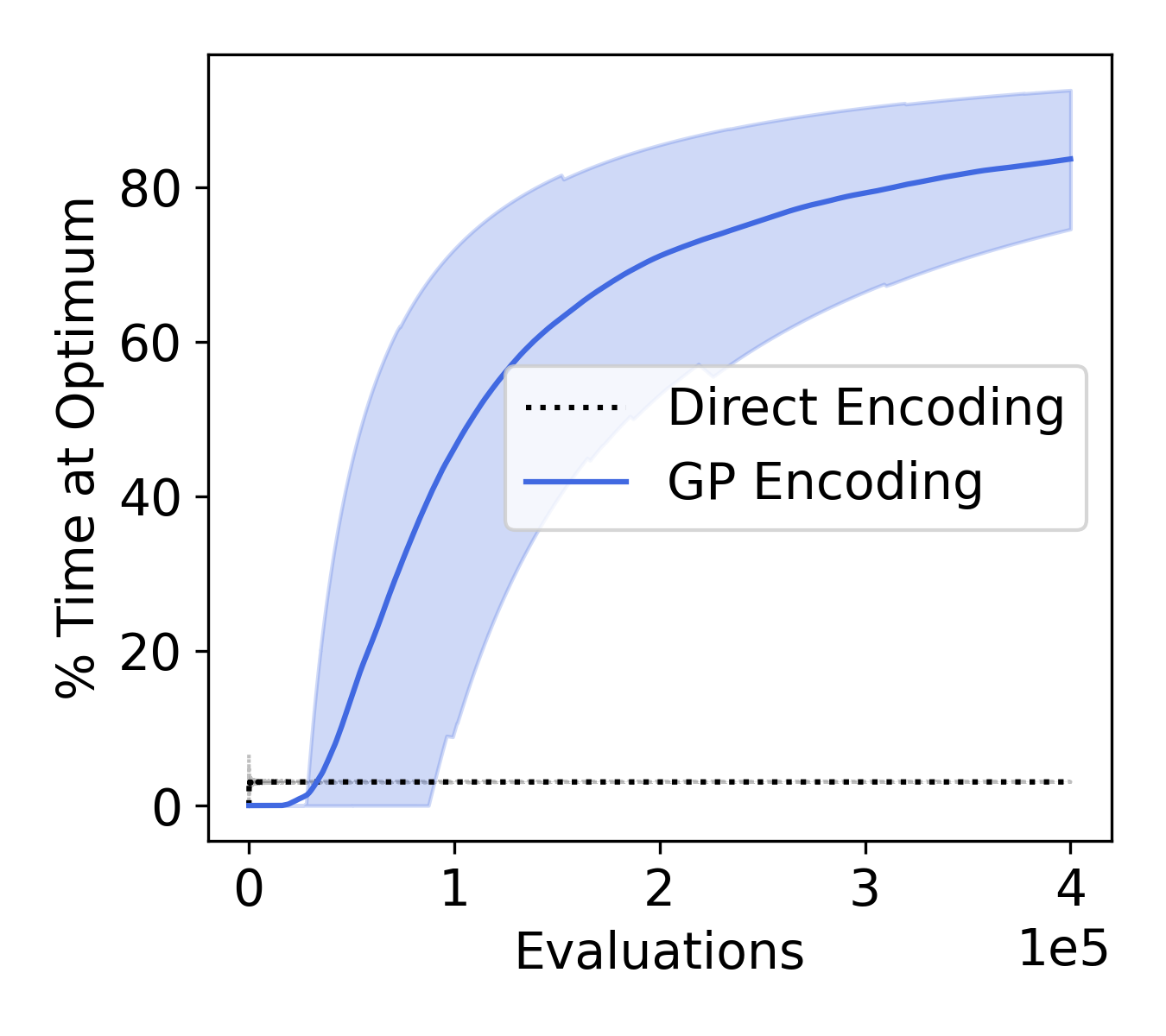}
\end{minipage}
\begin{minipage}{0.26\textwidth}
    \centering
    (d) RFC\\ \vspace{-2pt}
    \includegraphics[width=\linewidth]{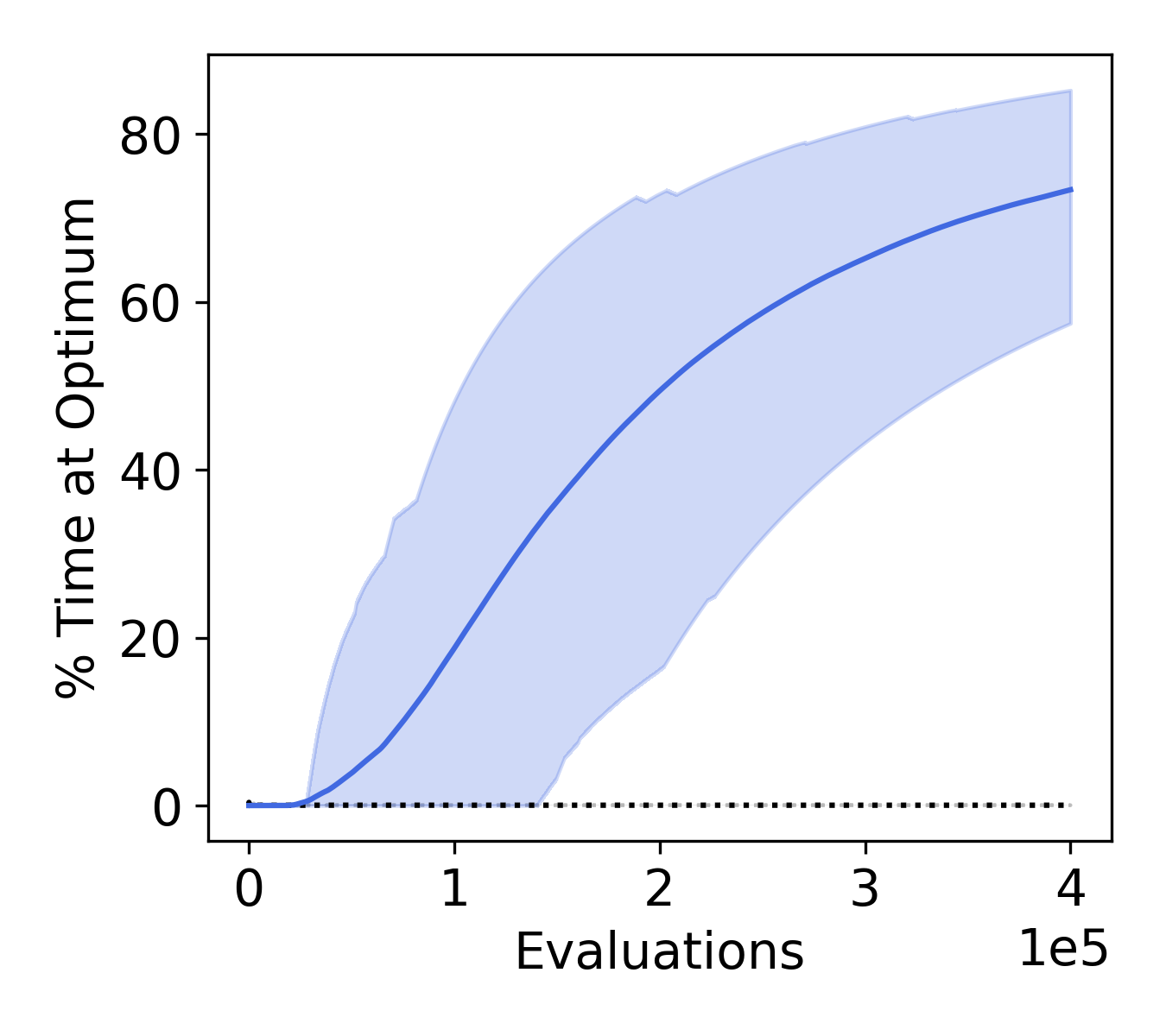}
\end{minipage}
\begin{minipage}{0.26\textwidth}
    \centering
    (e) LBAP\\ \vspace{-2pt}
    \includegraphics[width=\linewidth]{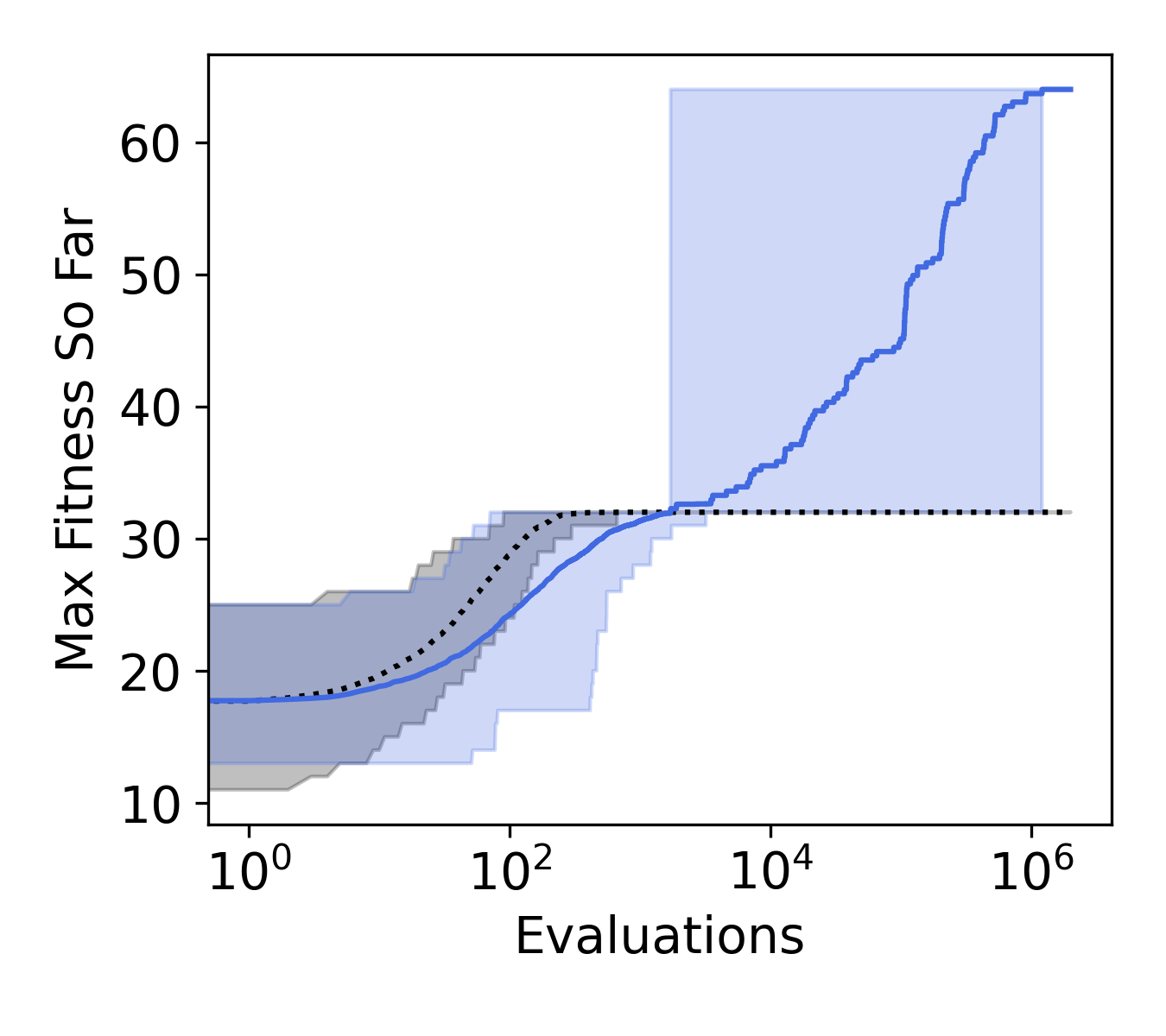}
\end{minipage} \vspace{-5pt}
    \caption{\emph{Adaptation and Convergence.}
    (a-b) Genotype templates for the encoding used in (a) Problems 1 and 2 and (b) Problem 3. The $\ba_*$, $\bb$, and $\bc$ are evolvable bit vectors with $\textrm{dim}(\bb) = \textrm{dim}(\bc) = n$.
    Distinct structure is learned in $\bb$ and $\bc$, whose coupling can then be exploited, temporally or in the phenotype space.
    (Note that these encodings are isomorphic to NNs with multiplicative units \cite{schmitt2002complexity} and weights in $\{-1,1\}$.)
    (c-d) Experimental results for Problems 1 and 2 with $n = 16, \lambda=2,$ and $L=10^5$.
    Consistent with its $O(1)$ adaptation complexity (Thm.~\ref{thm:deterministic_flip} and \ref{thm:random_flip}), the GP encoding spends an increasing proportion of time at optimal fitness; the direct encoding does not (mean, 90\% conf. over 100 trials).
    (e) Experimental results for Problem~3 with $n = 64, \lambda=1,$ and $R=10^5$, showing max, mean, and min over 100 trials. The GP encoding converges on all trials, while the direct encoding converges on none of them.
    Thus, with expressive encodings, even simple genotypes can lead to massive improvements.
    \label{fig:adaptation_and_convergence}}
\end{figure*}
For the expressive encoding, consider GP genotypes with the structure shown in Figure~\ref{fig:adaptation_and_convergence}(a),
where $\ba$, $\bb$, $\bc$ are bit vectors of length $\textrm{dim}(\ba) = L$, $\textrm{dim}(\bb) = \textrm{dim}(\bc) = n$.
So, the evolvable genome is defined by $x = (\ba, \bb, \bc)$, and has length $2n + L$.
The champion is initialized with random bits.
It turns out that GP results in a super-exponential speed-up: While the direct encoding takes $O(n\log n)$ to adapt to the new target, the GP encoding, initially ignorant of $\vy_1^*$ and $\vy_2^*$, spends only constant time.

\begin{thm}
\label{thm:p1_direct}
The (1+$\lambda$)-EA with direct encoding has $\Theta(n \log n)$ adaptation complexity on the deterministic flipping challenge.
\end{thm}
\begin{pfs}
This follows from a standard coupon collector argument \cite{doerr2020probabilistic}.
\qed \vspace{2pt}
\end{pfs}

\begin{thm}
\label{thm:deterministic_flip}
The (1+$\lambda$)-EA with GP encoding has $\ O(1)$ adaptation complexity on the deterministic flipping challenge.
\end{thm}
\begin{pfs}
Let $\dim(\ba) = L$.
Say $\parity(\ba) = 1$ and $f(\bb) > f(\bc)$.
Then, the algorithm only accepts mutations that move $\bb$ towards $\vy_1^*$.
Once $\bb = \vy_1^*$ and $\ba$ is mutated, $\bc$ will converge to $\vy_2^*$.
Then, if $\lambda=2$ and $L$ is large enough, the chance of making a mistake in $\bb$ or $\bc$ is so small that the expected time spent on repairs is constant.
\qed \vspace{7pt}
\end{pfs}

Thus, the expressive encoding yields asymptotically optimal adaptation complexity.
The reader may be concerned that this encoding takes longer to initially reach a global optimum than the direct encoding.
There are many ways to address this issue, e.g., by including mutations that enable $L$ to grow in size over time.
However, the next problem shows an even sharper advantage: Direct encoding spends exponential time away from the optimum, while the expressive encoding is asymptotically unaffected. 

\begin{problem}
\textbf{Random Flipping Challenge (RFC):}
Take any two target phenotypes $\vy_1^*, \vy_2^* \in \{0, 1\}^n$, where $\vy_2^*$ is the complement of $\vy_1^*$.
At each time $t$ the current target vector is selected to be $\vy_1^*$ or $\vy_2^*$ with 0.5 probability each.
The fitness is the number of bits in the phenotype that match the target. \vspace{5pt}
\end{problem}

This is the same as Problem~1, but with the target vector flipping randomly at each generation.

\begin{thm}
The (1+$\lambda$)-EA with direct encoding and $\lambda=2$ has $\Omega(2^{\nicefrac{n}{2}})$ adaptation complexity on the random flipping challenge.
\end{thm}
\begin{pfs}
Lower bound the hitting time of either target, using the fact that when within $\nicefrac{n}{4}$ bits of the target, the chance of moving towards it is less than half that of moving away. \qed
\end{pfs}

\begin{thm}
\label{thm:random_flip}
The (1+$\lambda$)-EA with GP encoding and $\lambda=2$ has $O(1)$ adaptation complexity on the random flipping challenge.
\end{thm}
\begin{pfs}
With $\lambda=2$, $L$ can be chosen large enough so the chance of moving towards the correct target is always more than twice that of moving away, which can be used to upper bound the hitting time.
Then, as in Theorem~\ref{thm:deterministic_flip}, once $\bb$ and $\bc$ have initially converged, the expected time spent on repairs is constant.
\qed \vspace{7pt}
\end{pfs}

The theoretical results for Problems~1 and~2 are validated experimentally in Figure~\ref{fig:adaptation_and_convergence}(c-d).
Consistent with its $O(1)$ adaptation complexity, the GP encoding spends an increasing overall percentage of time at optimal fitness.
The power of expressive encodings also manifests on static fitness functions, like the one in Problem~3. 

\begin{problem}
\textbf{Large Block Assembly Problem (LBAP):}
In this problem, there are two targets $\vy_1^*, \vy_2^* \in \{0, 1\}^\frac{n}{2}$ hidden somewhere amongst the $n$ bits at non-overlapping indices, with $|\vy_1^*|, |\vy_2^*| > 1$.
If the solution contains both targets the fitness is $n$, otherwise it is the maximum number of bits matched to either target. \vspace{5pt}
\end{problem}
That is, there are solutions of size $\Theta(n)$ in disjoint subspaces that must be discovered and combined.
This problem is challenging for a direct encoding: Once one target is found there is no fitness gradient until all remaining $\frac{n}{2}$ bits are matched, which takes exponential time to happen by chance.
Since the fitness function is fixed, the metric of interest is simply expected time to reach the optimum.

For this problem, there is no dynamism in the fitness function to provide exploratory power to the (1+$\lambda$)-EA, so some basic mechanisms must be added to prevent it from getting stuck.
Instead of simply accepting if the candidate has higher fitness, the champion is replaced if any of the following conditions are met:
\begin{itemize}
    \item \emph{Fitness}. The candidate has greater fitness.
    \item \emph{Diversity}. The candidate phenotype is further than Hamming distance one from the champion phenotype.
    \item \emph{Sparsity}. The candidate phenotype has equal fitness and fewer ones than the champion phenotype.
\end{itemize}
With these rules, the algorithm can still be subject to deception, so the champion is reinitialized after $R$ steps if it has not yet converged.
Call this adjusted algorithm (1+$\lambda$)-EA*.
The pitfalls of direct encoding are too great for these methods to help much in that case.

\begin{thm}
The (1+$\lambda$)-EA* with direct encoding and $\lambda=1$ converges in $\Omega(2^n)$ steps on the large block assembly problem.
\end{thm}
\begin{pfs}
Once one target is found, the algorithm is stuck, so both targets can only be found together by chance.
\qed \vspace{7pt}
\end{pfs}

However, using genotypes with the GP structure shown in Figure~\ref{fig:adaptation_and_convergence}(b), with $\dim(\ba_1) = \dim(\ba_2) = 1$, leads to tractability.
\begin{thm}
The (1+$\lambda$)-EA* with GP encoding and $\lambda=1$ converges in $O(n^{3 + o(1)})$ steps on the large block assembly problem.
\end{thm}
\begin{pfs}
Consider convergence paths where $\ba_1 = \ba_2 = \textbf{1}$ only occurs at the final step.
Compute the probability of such convergence, and use restarts to get the expected run time.
\qed \vspace{7pt}
\end{pfs}

These theoretical conclusions are demonstrated experimentally in
Figure~\ref{fig:adaptation_and_convergence}(e).
In 100 independent trials, with the direct encoding, the algorithm never reaches beyond half the optimal fitness; with GP encoding it always makes the final large jump before the 2M evaluation limit.

\subsection{Extensions}

The algorithms in this section worked with fixed GP structures, evolving the values within them.
This approach is sufficient to demonstrate the power and potential of expressive encodings; future work will analyze methods that evolve the structure of representations as well.
However, note that the above structures are each of a constant size: One could simply enumerate all such structures, trying the algorithm on each, and only pay a constant multiplicative cost, not affecting the asymptotic complexities.
Although this is a brute-force approach, it suggests that powerful structures may actually not be so difficult to find, and indeed meaningful structures are commonly found by existing GP methods \cite{schmidt2009distilling,shahrzad2015tackling}.

The problems in this section all required jumps of size $\Theta(n)$.
Prior work with direct encodings has sought to tackle larger and larger jumps, but they are still generally sublinear \cite{bambury2021generalized,dang2016escaping}.
The closest comparison with SGOs \cite{watson2007building} had jumps of size $O(\sqrt{n})$, but required a representation that was a priori well-aligned for two-point crossover, and an island model with a number of islands dependent on the $n$.
In contrast, with an expressive encoding, successful jumps of size $\Theta(n)$ can be achieved with only single-point mutation in a (1+1 or 2)-EA, and simple sparsity and diversity methods.

The diversity mechanism used for Problem 3 could also be used to generalize the first two problems to any two target vectors.
This simple diversity mechanism is an instance of a behavior domination approach \cite{meyerson2017discovering}, in that solutions at least a fixed distance from the current solution are non-dominated.

\section{Discussion and Future Work}
\label{sec:discussion_and_future_work}

The definitions and analysis of expressive encodings in this paper can be seen as a starting point for future work in several areas:

\vspace{3pt}\emph{Biological Interpretation.}
Because they can capture complex reproductive distributions, expressive encodings could in general be more accurate than direct encodings in representing complex models of biological evolution.
If genetic regulatory networks are universal function approximators (models of them often are, e.g., recurrent NNs, ODEs, and boolean networks \cite{crombach2008evolution,karlebach2008modelling,yaghoobi2012review}), then Theorem~\ref{thm:ufa} suggests how \emph{natural evolution} can become arbitrarily powerful over time.
In the expressive encodings with crossover constructions in Section~\ref{sec:sampling_and_universality}, the high level of shared structure across parents is also consistent with nature. Humans share >99\% of their DNA, and high proportions of DNA are even shared across species \cite{collins1997variations,hardison2003comparative,simakov2022deeply}.
This construction may partly explain why crossover is so prominent in biology: It works best when a large part of the genome is shared.
Another surprising result from Section~\ref{sec:adaptation_and_convergence} is that for expressive encodings in dynamic environments, generating multiple offspring per generation is critical to achieving stable performance.
This result makes biological sense as well: Even when the probability of a good offspring is high, if there is any chance that a bad one can set you back considerably due to latent nonlinearities in the genome, it is prudent to have multiple offspring.

\vspace{3pt}\emph{Theory.}
The theory in this paper did not use any deep EA theory methods, such as drift methods \cite{doerr2012multiplicative,doerr2013adaptive,lengler2020drift}.
Such methods could enable rapid extension of these results to other encodings, operators, and test domains.
To highlight the mechanisms that make expressive encodings powerful, the test domains were idealized to contain two complementary targets.
Future work can generalize this approach to more targets, and to compositions of targets that have not been seen before, akin to generalization in machine learning.
Section~\ref{sec:adaptation_and_convergence} demonstrated a type of stability, i.e., preventing catastrophic forgetting in particular cases.
How far can such results be generalized? For instance, can catastrophic forgetting be asymptotically avoided in individuals as complex as the ones constructed in Section~\ref{sec:sampling_and_universality}? What other kinds of stability are possible?

\vspace{3pt}\emph{Practice.}
In building practical applications, it is not clear whether the first step should use incrementally growing encodings like those in GP and NEAT \cite{stanley2002evolving}, or giant fixed structures with many evolvable parameters.
The EA community has tended to prefer incremental encodings, but deep learning has seen remarkable success via huge fixed structures with canonical form and learnable parameters.
Perhaps expressive encodings with a fixed structure would be an easier place to start, since understanding the behavior of the system may be easier without the variable of dynamic genotypic structure.
Also, one of the common motivations for indirect encodings is that the genotype can be smaller than the phenotype. However, in the constructions of Section~\ref{sec:sampling_and_universality} the genotype is generally \emph{much} larger than the phenotype, which suggests that directly evolving huge genotypes could be promising (similar to deep GA \cite{such2017deep}). 

\vspace{3pt}\emph{Open-ended Evolution.}
The power of expressive encodings comes from the ability of the transmission function to improve throughout evolution.
This phenomenon has been previously explored in GP \cite{altenberg1994evolution,hu2011robustness} and neuroevolution \cite{huizinga2018emergence,lehman2013evolvability}, and insights there should be useful in analyzing the behavior of expressive encodings more generally.
E.g., motivated by Evolvability-ES, a construction related to miracle jumps for directly-encoded NNs has been developed for i.i.d. perturbations \cite[Thm. S6.1]{gajewski2019evolvability}.
As a longer-term opportunity, the promise of expressive encodings to continually complexify and innovate upon the transmission functions indicates that they should be a good substrate for open-ended evolution \cite{hintze2019open,soros2014identifying,stanley2019open,stepney2021modelling,taylor2016open,wang2020enhanced}, and, likewise, insights from open-endedness could be useful in developing more practical SGO~+~Expressive Encoding algorithms.
Thus, the long-term ideal is an algorithm that never needs to be restarted: press `go' once and over time it becomes better and better at solving problems as they appear, accumulating problem-solving ability in the genotypes themselves.
Nature has provided an existence proof of such an algorithm.
For organisms to become boundlessly and efficiently more complex over time, the mapping from parents to offspring must become more complex over time.
Expressive encodings are an important piece of this puzzle.

\section{Conclusion}
\label{sec:conclusion}

This paper identifies expressivity as a	fundamental property that
makes encodings powerful. 
It allows local changes in the genotype to
have global effects in the phenotype, which is useful in making large
jumps in particular, and approximating arbitrary probability
distributions in general. 
Direct encodings are not generally
expressive, whereas genetic programming and neural network encodings
are. 
As demonstrated in three problems illustrating different
challenges, expressive encodings may bring striking improvements in adaptation and make intractable problems tractable.
We believe expressivity provides a productive perspective for further research on many aspects of evolutionary computation.



\begin{acks}
Thanks to the full Evolutionary AI research group at Cognizant for regular discussions throughout the course of the work.
Thanks to Babak Hodjat for ideas on experimental design.
Thanks also to Andrew Kelley for additional technical feedback.
\end{acks}

\bibliographystyle{ACM-Reference-Format}
\bibliography{expressive-encodings}

\clearpage

\appendix

\section{Proofs for Section~\ref{sec:sampling_and_universality}}
\label{sec:proofs_for_universality}

\vspace*{1ex}
\begin{manualtheorem}{4.1}
Genetic programming is an expressive encoding for uniform crossover, with complexity $O(mn - m\log \epsilon)$. \label{thm:gp_universal_crossover}
\end{manualtheorem}
\begin{proof}
Let $y'_1$, $y'_2$ be the phenotypes of the two parent arguments to the uniform crossover operator $g_c$.
Choose the parent genotypes $x'_1$, $x'_2$ to be the two programs shown in Figure~\ref{fig:gp_universal}(a).
These two programs differ only in their value of $\mathbf{a}$, so the child will be equivalent to its parents, except its value of $\mathbf{a}$ will be uniformly sampled from $\{0, 1\}^n$.

When interpreted as a binary integer, this value of $\mathbf{a}$ is an integer uniformly sampled from $[0, 2^L-1]$, where $L$ is the length of the vector $\mathbf{a}$.
Let each $\bt_i$ be a binary vector of length $L$, each of which can also be interpreted as an integer, with $\bt_1 < \bt_2 < \ldots < \bt_{m-1}$.
Now, the approach is to choose $L$ large enough and then choose the $\bt_i$'s so that each $\by_i$ is generated when the integer $\mathbf{a}$ falls in a given range, which occurs with probability within $\epsilon$ of $p_i$.

Let $L = \lceil \lg \frac{1}{\epsilon} \rceil + 2$ = $\Theta(-\log\epsilon)$.
Then, there are $2^L > \frac{2}{\epsilon}$ integers each sampled with probability $\frac{1}{2^L} < \frac{\epsilon}{2}$.
Let $\bt_1 = \lfloor p_1 L \rfloor$, $\bt_i= \lfloor L \sum_{j=1}^i p_{j} \rfloor \ \forall \ i \in [2, \ldots, m - 1]$.
Then, $\forall \ i \in [2, \ldots, m - 1]$
\begin{equation}
   p_i - \frac{\epsilon}{2} < \Pr[E_{GP}(g_c(x'_1, x'_2)) = y_i] \leq p_i. 
\end{equation}
In order to ensure that the parent genotypes generate their required phenotypes, the single integers $0\ldots0$ and $1\ldots1$ are assigned to $\by'_1$ and $\by'_2$.
The probabilities of sampling these are each less than $\frac{\epsilon}{2}$, and are subtracted from the probabilities for $\by_1$ and potentially $\by_m$, so that for $j \in \{1, m\}$,
\begin{equation}
  p_j - \epsilon < \Pr[E_{GP}(g_c(x'_1, x'_2)) = \by_j] \leq p_j.  
\end{equation}
In the parent genotypes, each conditional contains $O(L + n)$ bits (to encode $\bt_i$ and $\by_i$) and there are $m$ conditionals.
Since $L = O(-\log \epsilon)$, the required total genotype size is $O(-m\log \epsilon  + mn)$.
\end{proof}

\vspace*{1ex}
\begin{manualtheorem}{4.2}
Genetic programming is an expressive encoding for single-point mutation, with complexity $O(\frac{mn}{\epsilon})$.
\end{manualtheorem}
\begin{proof}
Let $y'$ be the phenotype of the single parent argument to the single-point mutation operator $g_m$.
Choose the parent genotype $x'$ to be the program shown in Figure~\ref{fig:gp_universal}(b), with each $\va_i$ a binary vector of zeros $[0, \ldots, 0]$.
Since $\parity(\va_i)$ is false for all $\va_i$ in $x'$, $E(x') = \by'$.
Our approach is to choose the length $L_i$ of each $\va_i$ so that the probability of selecting a bit to flip in $\va_i$ is approximately $p_i$.

When $g_m$ chooses a bit to flip, it selects the bit uniformly from the $\va_i$'s and the $\vy_i$'s.
The total size of the $\vy_i$'s is $mn$.
So, let
\begin{equation}
   \sum_{i=1}^m L_i = \bigg\lceil \frac{mn}{2\epsilon} \bigg\rceil.
\end{equation}
Then, the $L_i$'s can be apportioned proportionally to $p_i$ such that the chance of flipping a bit in $\ba_i$ out of all $\ba$'s is within $\frac{\epsilon}{2}$ of $p_i$.
The chance of instead flipping a bit in the $\by$'s is less than $\frac{\epsilon}{2}$, so the overall probability of flipping a bit in $\ba_i$ is within $\epsilon$ of $p_i$.

In this construction, the size of the $\va$'s dominate the $\by$'s, giving us a complexity of $O(\frac{mn}{\epsilon})$.
\end{proof}

Note that the parent in Figure~\ref{fig:gp_universal}(b) could have been more like Parent 1 in Figure~\ref{fig:gp_universal}(a), using a single long $\va$ with <.
Instead, the parity function `par' is used, because it demonstrates an alternative kind of construction, and because it is used again in Section~\ref{sec:adaptation_and_convergence}.

\vspace*{1ex}
\begin{manualtheorem}{4.3}
Feed-forward neural networks with sigmoid activation are an expressive encoding for uniform crossover, with complexity $O(mn - \log \epsilon)$.
\end{manualtheorem}
\begin{proof}

This construction is similar to the uniform crossover construction for genetic programming.
Let the parents $x'_1$, $x'_2$ be defined by the four-layer neural networks shown in Figure~\ref{fig:nn_universal}(a), where each internal node is followed by sigmoid activation, and all biases are 0.

The two parents differ only in the values of their second layer of weights.
Since all the input weights to the first layer are zero, the output of each node in the first layer is $0.5$.
Then, since each power of two is included with probability $\nicefrac{1}{2}$, uniform crossover makes the input to the bottleneck equal 0.5 times an integer sampled uniformly from $[0, 2^L-1]$.
Each such value uniquely defines the output of the bottleneck neuron.

Similarly to the GP case, a set of thresholds is then created.
First, $c_1$ is set to amplify the differences between successive outputs from the bottleneck, and the biases of the threshold units are then set so that they output nearly one if the threshold is met and nearly zero otherwise.
For brevity, we say that the unit ``fires'' if its output is nearly one.

There is a switch neuron for each desired $\vy_i$.
By setting $c_2$ as large as we want, this switch will fire only if its lower threshold unit fires, but its upper one does not.
Since the thresholds are monotonically increasing, exactly one switch unit will fire.

Finally, the weights from the switch for $\vy_i$ to the output of the whole network are $c_3(2\vy_i - 1)$, with $c_3$ arbitrarily large.
So, if the switch from $\vy_i$ fires, the input to the $j$th output node will be negative if the $j$the element of $\vy_i$ is 0, and positive if the $j$the element of $\vy_i$ is 1.
Therefore, after squashing through the final sigmoid, the output of the entire network will be nearly $\vy_i$, which after rounding yields $\vy_i$ exactly (see Equation~\ref{eq:nn_encoding}).
That is, the NN outputs the binary vector $\vy_1$.

Note that similar considerations as in the case of GP can be taken into account to ensure that $E(x'_1) = \by'_1$ and $E(x'_2) = \by'_2$.

As in the case of GP, it is necessary that $L = \Theta(-\lg \epsilon)$.
There are $\Theta(-\lg \epsilon)$ weights in the first two layers, $O(m)$ in the next two layers, and $O(mn)$ in the final layer.
So, in terms of number of weights, the complexity of this construction is $O(mn - \log \epsilon)$.
The apparent improvement over the GP construction comes from the fact that real-numbered weights were used instead of bits.
\end{proof}

\vspace*{1ex}
\begin{manualtheorem}{4.4}
Feed-forward neural networks with sigmoid activation are an expressive encoding for single-point mutation, with complexity $O(\frac{mn}{\epsilon})$.
\end{manualtheorem}
\begin{proof}
Take the single parent $x'$ for mutation to be Parent 2 in Figure~\ref{fig:nn_universal}(a).
In fact, in this case, the exact weights in the first two layers of $x'$ do not matter, as long as the weights in the second layer are non-zero.
A weight in the first two layers is selected for mutation with probability of at least $1 - \nicefrac{\epsilon}{2}$, by choosing a large enough $L = \Theta(\frac{mn}{\epsilon})$.
Now, no matter what the weights are in the first two layers of $x'$, the stochastic process of (1) selecting one of these weights, and (2) adding Gaussian noise to this selected weight yields \emph{some} continuous distribution over the output of the bottleneck unit.
Similar to the case of crossover, the biases of the threshold units can be set to partition the output distribution of the threshold unit appropriately, again setting $c_2$ and $c_3$ high enough to make the output of each unit nearly one or nearly zero.
It is also possible to include threshold units for the case of no mutation to ensure that $E(x') = \by'$.
Similar to the construction for GP with mutation, the complexity of this construction is dominated by $L$, i.e., the overall complexity is $O(\frac{mn}{\epsilon})$.
\end{proof}

\vspace*{1ex}
\begin{manualtheorem}{4.5}
\label{thm:ufa}
$E_\Omega$ is an expressive encoding for uniform crossover.
\end{manualtheorem}
\begin{proof}
As in the previous constructions for uniform crossover, choose $L = \Theta(-\lg \epsilon)$.
Let $x'_1 = \omega([0,\ldots,0])$ and $x'_2 = \omega([1,\ldots,1])$.
Again, since the parents differ only in their values of $\va$, the child will be of the form $\omega(\mathbf{b})$ where $\mathbf{b}$ is drawn uniformly from $\{0, 1\}^L$.
It is clear from previous constructions that values in $\{0, 1\}^L$ can be apportioned to assign accurate-enough probability to each $y_i$, and $\omega$ can thus be chosen so that it assigns probability in this way.
That is, $\{0, 1\}^L$ is partitioned into subsets $S_i$ so that $\nicefrac{\lvert S_i \rvert}{2^L} \approx p_i$, and $\omega$ is chosen so that if $\mathbf{b} \in S_i$ then $\omega(\mathbf{b}) = \by_i$, while also ensuring that $\omega([0,\ldots,0]) = \by'_1$ and $\omega([1,\ldots,1]) = \by'_2$.
\end{proof}

\vspace*{1ex}
\begin{manualtheorem}{4.6}
Direct encoding of feed-forward neural networks with sigmoid activation is an expressive encoding for uniform crossover.
\end{manualtheorem}
\begin{proof}
The phenotype is directly encoded as a neural network. That is, the phenotype space $H$ consists of functions $H$, where each $h \in H$ can be represented as a neural network with $h: \mathbb{R}^{n_\mathrm{in}} \to \mathbb{R}^{n_\mathrm{out}}$.
Given a desired set of such functions $h_i \ (i = 1, \ldots, m)$ and associated probabilities $p_i$ with $\sum_i p_i = 1$, the goal is to find directly-encoded parents $x'_1 = h'_1$ and $x'_2 = h'_2$ whose crossover results in the desired distribution.

A similar approach can be taken as with previous constructions, but with direct encodings of neural networks, the source of randomness can be placed in the biases of nodes whose input weights are all zero, so that they effectively serve as auxiliary inputs to the model.

Let $h'_1$ and $h'_2$ be the two parents shown in Figure~\ref{fig:nn_universal}(b).
Let the first hidden layer of both $h'_1$ and $h'_2$ have $n_{in} + L$ units $u_1,\ldots,u_{n_{in} + L}$.
For $u_j \in u_1,\ldots,u_{n_{in}}$ let the incoming weights to $u_j$ all be zero except the weight from the $j$th input, which is set to 1.
Let the biases of $u_1,\ldots,u_{n_{in}}$ all be 0.
For $u_j \in u_{n_{in}+1},\ldots,u_L$ let the incoming weights to $u_j$ all be zero.
Let the biases of $u_{n_{in}+1},\ldots,u_L$ all be 0 in $h'_1$ and all be 1 in $h'_2$.
Let the remainder of $h'_1$ and $h'_2$, denoted $h'':\mathbb{R}^{n_{in} + L} \to \mathbb{R}^{n_{out}}$, be shared.

Then, a child generated by $g_c(h'_1, h'_2)$ will be equivalent to its parents, except the biases of $u_{n_{in}+1},\ldots,u_L$ will be sampled uniformly from $\{0, 1\}^L$.
Choosing $L = \Theta(-\lg \epsilon)$, since feed-forward NNs are universal function approximators, $h''$ can again be selected so that each $h_i$ is sampled with approximately the desired probability---that is, the sampled biases tell $h''$ which function to compute.
Importantly, the sigmoid activation after the first layer of $h'_1$ and $h'_2$ does not lose any information about the input, since it is continuous and strictly monotonic, so the input to $h''$ still uniquely identifies the input to the whole network.
\end{proof}

\section{The (1+$\lambda$)-Evolutionary Algorithm}
\label{sec:algorithm1}

Algorithm~\ref{alg:1_plus_lambda_ea} provides pseudocode for the algorithm used in Section~\ref{sec:adaptation_and_convergence}.
$E$ is the encoding and $f$ is the fitness function.

\begin{algorithm}
\caption{(1+$\lambda$)-EA}\label{alg:1_plus_lambda_ea}
\begin{algorithmic}
\State $x_o \gets$ random initial genotype \Comment{Initialize champion}
\While{not done}
\For{$i \in 1, \ldots, \lambda$}
    \State $x_i \gets \textrm{mutate}(x_o)$ \Comment{Generate candidates}
\EndFor
\For{$i \in 1, \ldots, \lambda$}
    \If{$f(E(x_i)) > f(E(x_o))$}
        \State $x_o \gets x_i$ \Comment{Replace Champion}
    \EndIf
\EndFor
\EndWhile
\end{algorithmic}
\end{algorithm}

\section{Proofs for Section~\ref{sec:adaptation_and_convergence}}
\label{sec:proofs_for_convergence}

The proofs in this Section consider the (1+$\lambda$)-EA (Algorithm~\ref{alg:1_plus_lambda_ea}), with $\lambda \in \{1, 2\}$.

\vspace*{1ex}
\begin{manualtheorem}{5.1}
The (1+$\lambda$)-EA with direct encoding has $\Theta(n \log n)$ adaptation complexity on the deterministic flipping challenge.
\end{manualtheorem}
\begin{proof}
This follows from a standard coupon collector argument \cite{doerr2020probabilistic}, where it takes $\Theta(n \log n)$ steps to move to a point $\Theta(n)$ bits away.
Increasing $\lambda$ to 2 does not improve the complexity.
\end{proof}

\vspace*{1ex}
\begin{manualtheorem}{5.2}
\label{thm:deterministic_flip}
The (1+$\lambda$)-EA with GP encoding has $O(1)$ adaptation complexity on the deterministic flipping challenge.
\end{manualtheorem}
\begin{proof}
W.l.o.g. assume $\vy_1^*$ is the vector of ones and $\vy_2^*$ is the vector of zeros.
The crux of the proof is showing that, once the algorithm converges initially, if $\lambda$ is at least 2, it is possible to pump $L$ large enough so that the expected time spent recovering from forgetting is constant.

\paragraph{Initial Convergence}
Let $x$ be initialized uniformly at random. Say the initial target is the ones vector $\mathbf{1}$.
Let $|\cdot|$ denote the 1-norm, i.e., number of 1's in the vector.

Case 1: Suppose $\text{par}(\ba)$ and $|\bb| > |\bc|$. Then the only accepted children are ones that flip a 0 to a 1 in $\bb$.
In this case, $\bb$ will converge to $\mathbf{1}$ in $O((L+n)\log n)$ steps, by a classic coupon collector argument.

Now the target flips to the zeros vector $\textbf{0}$.
Suppose $L = \omega(n)$. Then, in a constant number of steps a bit of $\ba$ is flipped, thereby returning $\bc$.
(If, in the meantime a bit of $\bb$ is accidentally flipped, the expected cost to repair this mistake can also be constant, as shown in the next section).
Since $|\bc| < |\bb|$, now the only accepted children will be ones that flip a 1 to a 0 in $\bc$.

Case 2: Suppose $\neg \text{par}(\ba)$ and $|\bc| > |\bb|$. By symmetry, the same result is obtained as in Case 1.

There is a $\frac{1}{2}$ probability that a random initialization results in Case 1 or Case 2.
If not, Case 1 or 2 can be established by either flipping a bit of $\ba$ or flipping bits of whichever of $\bb$ or $\bc$ is being returned until it has more 1's than its counterpart.

The expected time to flip a bit of $\ba$ is $\frac{2n + L}{L}$, which is $O(1)$ when $L = \omega(n)$, so this initial time to get to Case 1 or 2 is negligible. 
The overall expected time to convergence is then $O((L+n)\log n)$.

\paragraph{Repeated Convergence}
After initial convergence, w.l.o.g. suppose $|\bb| = n$, $|\bc| = 0$, and $\text{par}(\ba) = 1$, so the current target is $\mathbf{0}$, and the fitness of the champion $x_i$ is now $f(x_i) = 0$.

A new candidate $f(x'_i)$ can be generated in three ways:

Case 1: flip a bit of $\ba$, so $f(x'_i) = n$ and the process is done.

Case 2: flip a bit of $\bb$, so $f(x'_i) = 1 > f(x_i)$.

Case 3: flip a bit of $\bc$, so $f(x'_i) = 0 = f(x_i)$.

Only Case 2 is of concern, since in this case an important bit of $\bb$ can be forgotten when the new candidate is accepted.
In this case, up to $n-1$ mistakes can be made in $\bb$ before a bit is finally flipped in $\ba$.
The chance of making $k$ mistakes in a row vanishes quickly, but even in the worst case, it would only take $O((n+L)\log n)$ steps to make the repairs.
So, the expected time needed to make repairs is less than the probability of making at least one mistake times the cost of making repairs, i.e.,
$\E$(repairs) = $\Pr$(mistake) $\cdot \ O((n+L)\log n)$.

Suppose $\lambda=1$. Then,
\begin{equation}
\text{$\Pr$(mistake)} = \frac{n}{2n + L}, \textrm{~and}
\end{equation}
\begin{equation}
\E\text{(repairs)} =\\ O\bigg(\frac{n}{n + L}\bigg) \cdot O((n+L)\log n) = O(n \log n).    
\end{equation}
That is, $O(n \log n)$ steps are spent on repairs every time the target flips, which is no better than the direct encoding case.

Suppose instead $\lambda=2$. Then,
\begin{equation}
   \Pr\text{(mistake)} = \bigg(\frac{n}{2n + L}\bigg)^2, \textrm{~and}
\end{equation}
\begin{equation}
\E\text{(repairs)}  =\\ O\bigg(\bigg(\frac{n}{n + L}\bigg)^2\bigg) \cdot O((n+L)\log n) = O\bigg(\frac{n^2 \log n}{n + L}\bigg).
\end{equation}
So, choosing $L = \omega(n^2 \log n)$ leads to
\begin{equation}
    \E\text{(repairs)} = O\bigg(\frac{n^2 \log n}{n + L}\bigg) = O(1).
\end{equation}

Thus, both the expected time to hit Case 1 and the expected time to make any necessary repairs are constant.
\end{proof}

\vspace*{1ex}
\begin{manualtheorem}{5.3}
The (1+$\lambda$)-EA with direct encoding and $\lambda=2$ has $\Omega(2^{\nicefrac{n}{2}})$ adaptation complexity on the random flipping challenge.
\end{manualtheorem}
\begin{proof}
W.l.o.g., suppose $\vy_1^*$ and $\vy_2^*$ are the vectors of all zeros and all ones.
Suppose the champion $x$ contains $k$ ones at time $t$, and generates candidate $x'$ at time $t+1$.
If the target at time $t+1$ is all-ones, and a zero in $x$ is flipped to a one, then the change is accepted
This flip happens with probability $\frac{1}{2} \cdot (1 - (\frac{k}{n})^2)$.
If the target at time $t+1$ is all-zeros, and a one in $x$ is flipped to a zero, then the change is accepted.
This flip happens with probability $\frac{1}{2} \cdot (1 - (\frac{n-k}{n})^2)$.
The $x$ is kept with the remaining probability, which is greater than $\frac{1}{2}$.

Now, if $k < (1 - (\nicefrac{\sqrt{12}}{2} - 1))n < \nicefrac{n}{4}$, then the chance of moving towards the target is less than half that of moving away.
Note that with $\lambda=1$, this sufficient $k$ is even higher, i.e., $\frac{1}{3}$.
So, suppose $k = \lfloor \frac{n}{4} \rfloor$, and let $h^0_i$ be the expected hitting time of reaching all zeros starting with $i$ ones.
Clearly, $h^0_k \geq 1 + h^0_{k-1}$.
Suppose $i \leq k$, and $h^0_{i+1} \geq c_{i+1} + h^0_i$ for some $c_{i+1} \in \mathbb{R}$.
Now,
\begin{align}
    h^0_i &\geq 1 + \frac{1}{6}h^0_{i-1} + \frac{1}{2}h^0_i + \frac{1}{3}h^0_{i+1}\\   
    \implies h^0_i &\geq 1 + \frac{1}{6}h^0_{i-1} + \frac{1}{2}h^0_i + \frac{1}{3}(c_{i+1} + h^0_i)\\
    \implies h^0_i &\geq 6 + 4c_{i+1} + h^0_{i-1} = c_i + h^0_{i-1}.
\end{align}
Now, $c_k = 1$, so $c_1 = \Omega(4^{\nicefrac{n}{4}}) \implies h^0_1 = \Omega(4^{\nicefrac{n}{4}}) = \Omega(2^{\nicefrac{n}{2}})$.
That is, the pull towards the center is so strong that even when the champion is one bit away from a target, the expected time to a target is $\Omega(2^{\nicefrac{n}{2}})$.
When at a target, the chance of moving away from it is constant, so this time to again reach a target dominates the adaptation complexity.
\end{proof}

\vspace*{1ex}
\begin{manualtheorem}{5.4}
The (1+$\lambda$)-EA with GP encoding and $\lambda=2$ has $O(1)$ adaptation complexity on the random flipping challenge.
\end{manualtheorem}
\begin{proof}
W.l.o.g. assume $\vy_1^*$ is the vector of ones, $\vy_2^*$ is the vector of zeros, and suppose that $|\bb| > |\bc| + 1$ and $\parity(\ba) = 1$ in $x^0$.
The question is how long it takes for $\bb$ to become all ones.
The $|\bb|$ is improved if the target is all-ones a zero is flipped in $|\bb|$ and the change is accepted; call this a \emph{fix}.
The $|\bb|$ is hurt if the target is all-zeros and a one is flipped in $|\bb|$ and the change is accepted; call this a \emph{break}.
Let $\lambda = 2$.
A break occurs if both candidates make a mistake, or one does and the other is neutral (i.e., it flips a bit in $\bc$).
So,
\begin{equation}
    \Pr(\mathrm{break}) = \frac{1}{2} \Bigg( \frac{|\bb|^2 + 2|\bb|n}{(2n + L)^2}  \Bigg).
\end{equation}
The probability of a fix is the chance that neither does not make an improvement:
\begin{equation}
    \Pr(\mathrm{fix}) = \frac{1}{2} \Bigg(1 - \Bigg( \frac{2n + L - (n - |\bb|)}{2n+L} \Bigg)^2 \Bigg).
\end{equation}
Of interest is the ratio $\nicefrac{\Pr(\mathrm{fix})}{\Pr(\mathrm{break})}$.
Choose $L > 3n^2  - 5n + \frac{3}{2}$.
Then, with some algebra, it can be seen that $\nicefrac{\Pr(\mathrm{fix})}{\Pr(\mathrm{break})} > 2$ for all $|\bb| < n$.

Now, let $h_i$ be the expected hitting time of $\bb$ to reach all-ones starting from $i$ ones.
For brevity, say $\Pr(\mathrm{fix})$ given $i$ ones is $s_i$.
Then,
\begin{equation}
    h_0 = 1  + s_0h_1 + (1 - s_0)h_0 \implies h_0 = \frac{1}{s_0} + h_1 = c_0 + h_1.
\end{equation}
Suppose $h_{i-1} \leq  c_{i-1} + h_i$. Then,
\begin{align}
    h_i &\leq 1 + \frac{s_i}{2}h_{i-1} + s_ih_{i+1} + \bigg(1 - \frac{3s_i}{2}\bigg)h_i\\
    \implies h_i &\leq \frac{1}{s_i} + \frac{1}{2}c_{i-1} + h_{i+1} = c_i + h_{i+1}. \textrm{~Then,}
\end{align}
\begin{align}
    h_0 &= \bigg(\frac{2n + L}{n}\bigg) + \sum_{i=1}^n c_i \leq \bigg(\frac{2n + L}{n}\bigg) + \sum_{i=1}^n \frac{1}{s_i} + \frac{1}{2}c_{i-1}\\
                     &\leq \bigg(\frac{2n + L}{n}\bigg) + \sum_{i=1}^n \frac{2n + L}{n - i} + \frac{1}{2^i}\bigg(\frac{2n + L}{n}\bigg)\\
    &= O\bigg(\frac{L}{n}\bigg) + O((n + L)\log n) = O((n + L)\log n).
\end{align}
So, the complexity for $|\bb|$ to converge from any starting point is the same as in Theorem~\ref{thm:deterministic_flip}, and, as in Theorem~\ref{thm:deterministic_flip}, if $|\bb| = n$,
\begin{equation}
   \Pr\text{(mistake)} = O\bigg(\bigg(\frac{n}{n + L}\bigg)^2\bigg), \textrm{~and}
\end{equation}
\begin{equation}
\E\text{(repairs)} =\\ O\bigg(\bigg(\frac{n}{n + L}\bigg)^2\bigg) \cdot O((n+L)\log n) = O\bigg(\frac{n^2 \log n}{n + L}\bigg).
\end{equation}
So, again choosing $L = \omega(n^2 \log n)$ leads to
\begin{equation}
    \E\text{(repairs)} = O\bigg(\frac{n^2 \log n}{n + L}\bigg) = O(1),
\end{equation}
and if $\bb$ and $\bc$ are both converged, then there is a constant chance of sampling the correct target at each step.
\end{proof}

\begin{manualtheorem}{5.5}
The (1+$\lambda$)-EA* with direct encoding and $\lambda=1$ converges in $\Omega(2^n)$ steps on the large block assembly problem.
\end{manualtheorem}
\begin{proof}
Since only a single bit is flipped, the diversity method has no effect on this problem.
The sparsity method is not useful either: It only biases the search towards 0's, which is not helpful in general; in the worst case, both hidden targets are all 1's.
In this case, the algorithm cannot make any progress once one of the targets is found, unless the state is already only one bit away from reaching the second target.
So, the best strategy is to restart every iteration, i.e., set $R = 1$, leading to a convergence time of $\Theta(2^n)$.
\end{proof}

\begin{manualtheorem}{5.6}
The (1+$\lambda$)-EA* with GP encoding and $\lambda=1$ converges in $O(n^{3 + o(1)})$ steps on the large block assembly problem.
\end{manualtheorem}
\begin{proof}
Call the two hidden targets $\bt_1$ and $\bt_2$.
W.l.o.g., suppose $\bt_1$ occurs in the first $\frac{n}{2}$ bits of a solution, and $\bt_2$ occurs in the last $\frac{n}{2}$ bits.
Let $[\bb_1, \bb_2] = \bb$ and $[\bc_1, \bc_2] = \bc$, where $\dim(\bb_1) = \dim(\bb_2) = \dim(\bc_1) = \dim(\bc_2) = \frac{n}{2}$.
Since $\dim(\ba_1) = \dim(\ba_2) = 1$, refer to the scalar contents of $\ba_1$ and $\ba_2$ as $a_1$ and $a_2$, respectively.

Upon random initialization, there is a constant probability that $|\bb_1 - \bt_1| < |\bb_2 - \bt_2|$, $|\bc_2 - \bt_2| < |\bc_1 - \bt_1|$, and either $a_1 = 0$ or $a_2 = 0$.
We are interested in how long it takes to reach the state $\bb_1 = \bt_1$, $\bb_2 = \textbf{0}$, $\bc_1 = \textbf{0}$, $\bc_2 = \bt_2$, and $a_1 = a_2 = 1$, since this is a state of maximal fitness.

However, if a state with $a_1 = a_2 = 1$ is reached before $\bb = [\bt_1, \textbf{0}]$ and $\bc = [\textbf{0}, \bt_2]$, then the states in $\bb$ and $\bc$ can be pulled in the wrong direction, due to the interaction between $\bb$ and $\bc$ caused by `$\oplus$'.

So, suppose that the algorithm reaches a state with $\bb_1 = \bt_1$, $\bb_2 = \textbf{0}$, $\bc_1 = \textbf{0}$, and $\bc_2 = \bt_2$, before ever reaching a state with $a_1 = a_2 = 1$.
Consider, then, the remaining three cases, which the algorithm alternates between as it converges:

Case 1: $a_1 = 1$ and $a_2 = 0$.
In this case, a new solution is kept if (1) a bit in $\bb_1$ is flipped that makes $\bb_1$ closer to $\bt_1$ (due to the \emph{fitness} condition, since the fitness will increase), if (2) a bit in $\bb_2$ is flipped from a 1 to a 0 (due to the \emph{sparsity} condition, since the sparsity will increase), or if (3) $a_1$ is flipped, which results in Case 2 (due to the \emph{diversity} condition, since disabling $\bb$ flips more than one bit in $\by$).

Case 2: $a_1 = 0$ and $a_2 = 0$.
In this case, a new solution is kept if $a_1$ or $a_2$ is flipped, resulting in Case 1 and Case 3, respectively (diversity condition).
No other flips affect the phenotype.

Case 3: $a_1 = 0$ and $a_2 = 1$.
In this case, a new solution is kept if (1) a bit in $\bc_2$ is flipped that makes $\bc_2$ closer to $\bt_2$ (fitness condition), if (2) a bit in $\bc_1$ is flipped from a 1 to a 0 (sparsity condition), or if (3) $a_2$ is flipped, resulting in Case 2 (diversity condition).



In expectation, half of the bits in $\bb$ and $\bc$ are correct upon initialization, and there are $2n$ possible bits that can be flipped without moving to a different Case.
So, since single-point mutation is used, the convergence of $\bb$ and $\bc$ are each equivalent to a coupon collector problem where there are $2n$ unique coupons of which a particular $\frac{n}{2}$ must be collected \cite{doerr2016impact}.
The expected convergence time of this process is
\begin{align}
    2nH_{\nicefrac{n}{2}} &\leq 2n\Big(\log \frac{n}{2} + 1\Big) < 2n(\log n + 1),
\end{align}
where $H_k$ is the $k$th harmonic number.

Each time Case 1 or 3 is visited, the process remains there for at least $2n$ steps in expectation, before moving to Case 2, from which there is a fifty-fifty chance of next moving to Case 1 or Case 3.
Let $V_1$ and $V_3$ be the number of visits before convergence for Case 1 and Case 3, resp.
Each case must be visited $H_{\nicefrac{n}{2}}$ times to converge in expectation, and the number of visits to either case is sampled from a binomial distribution $B(t, \frac{1}{2}, \frac{1}{2})$, where $t$ is the number of total visits $V_1 + V_3$.
Of interest is how many trials $t$ the binomial distribution needs to ensure that $\min(V_1, V_3) \geq H_{\nicefrac{n}{2}}$.
This can be achieved by finding $t$ so that the mean minus the standard deviation of the distribution is at least $H_{\nicefrac{n}{2}}$:
\begin{align}
    \mu\Big(B\Big(t, \frac{1}{2}, \frac{1}{2}\Big)\Big) - \sigma\Big(B\Big(t, \frac{1}{2}, \frac{1}{2}\Big)\Big) = \frac{t}{2} - \sqrt{\frac{t}{4}} \geq H_{\nicefrac{n}{2}}\\
    \implies t - \sqrt{t} - 2 H_{\nicefrac{n}{2}} \geq 0\\
    \implies t \geq 2H_{\nicefrac{n}{2}} + \frac{1}{2}\sqrt{8H_{\nicefrac{n}{2}} + 1} + \frac{1}{2} \\
    \implies t \geq 2H_{\nicefrac{n}{2}} + O\big(\sqrt{\log n}\big).
\end{align}
That is, in expectation, after $t$ total visits, the case visited least is visited $H_{\nicefrac{n}{2}}$ times and the case visited most is visited $H_{\nicefrac{n}{2}} + O(\sqrt{\log n})$ times.


Overall, the state converges to $\bb_1 = \bt_1$, $\bb_2 = \textbf{0}$, $\bc_1 = \textbf{0}$ and $\bc_2 = \bt_2$ after spending $O(n \log n)$ steps in Case 1 and $O(n \log n)$ steps in Case 3.
By symmetry, $O(n \log n)$ time is also spent in Case 2.
Since $O(n \log n)$ steps are spent in each case, the total time is also $O(n \log n)$, at which point $O(n)$ additional steps are needed to reach $a_1 = 1$ and $a_2 = 1$, completing convergence.

What is the chance that $a_1 = 1$ and $a_2 = 1$ is not reached until both $\bb$ and $\bc$ have converged?
Suppose $\max(V_1, V_3) = V_1$.
Then, for Case 1, the chance that $a_2$ is not flipped before $\bb$ converges is no less than
\begin{align}
    \Big(1 - \frac{1}{2n + 2}\Big) ^ {2n (H_{\nicefrac{n}{2}} + O(\sqrt{\log n}))} &= \Big(\frac{1}{e}\Big)^{H_{\nicefrac{n}{2}} + O(\sqrt{\log n})}\\
    &> \Big(\frac{1}{e}\Big)^{\log n + 1 + O(\sqrt{\log n})}\\
    &= \frac{1}{n} \cdot \frac{1}{e} \cdot \Big(\frac{1}{e}\Big)^{O(\sqrt{\log n})}\\
    &= \frac{1}{n} \cdot \frac{1}{e} \cdot \frac{1}{n^\epsilon} = \Omega\bigg(\frac{1}{n^{1 + \epsilon}}\bigg),
\end{align}
for any $\epsilon > 0$.
The last step uses from the fact that $e^{O(\sqrt{\log n})} = o(n^\epsilon)$ for any $\epsilon > 0$, which can be seen from the following \cite{vonbrand2019}:
\begin{align}
    \lim_{n \to \infty} \frac{e^{\alpha \sqrt{\log n}}}{\beta n^\epsilon} &= \exp \bigg( \log \lim_{n \to \infty} \frac{e^{\alpha \sqrt{\log n}}}{\beta n^\epsilon}\bigg) \\
    &= \exp\big( \lim_{n \to \infty} \alpha \sqrt{\log n} - \epsilon \log n - \log \beta \big) \\
    &= \exp(- \infty) = 0,
\end{align}
for any constants $\alpha >0$, $\beta > 0$.

By symmetry, there is the same lower bound on the probability that $\bc$ converges before $a_1$ is flipped in Case 3.
So, the probability that neither event occurs is
\begin{equation}
    \Omega\bigg(\bigg(\frac{1}{n^{1 + \epsilon}}\bigg)^2\bigg) = \Omega\bigg(\bigg(\frac{1}{n^{1 + o(1)}}\bigg)^2\bigg) = \Omega\bigg(\frac{1}{n^{2 + o(1)}}\bigg)
\end{equation}

Therefore, one can choose a restart threshold $R = \Theta(n \log n)$, with convergence expected in the above manner after $O\big(n^{2 + o(1)}\big)$ restarts, yielding an overall convergence time of $O\big(n^{3 + o(1)} \log n\big) = O\big(n^{3 + o(1)}\big)$.
\end{proof}


\end{document}